\documentclass[10pt,journal,compsoc]{IEEEtran}
\usepackage[T1]{fontenc}
\usepackage{amsmath,amssymb}
\usepackage[nocompress]{cite}
\usepackage{algpseudocode,booktabs,float,graphicx,multirow,placeins}
\usepackage{capt-of,adjustbox,xurl,xspace}
\usepackage[hidelinks]{hyperref}
\newcommand{\system}{AgentTriage\xspace}
\newcommand{\DiagRepositoryURL}{\url{https://github.com/AsnGs/agenttriage-open-source}}
\newcounter{algorithm}
\newfloat{diagnosisalgorithm}{tbp}{loa}

\newcommand{\DiagTargetCount}{5}

\newcommand{\DiagCollectedCases}{126}
\newcommand{\DiagResolvedCases}{120}

\newcommand{\DiagReferenceCoveragePct}{95.2}

\newcommand{\DiagMinimumCasesPerTarget}{24}
\newcommand{\DiagWeakLayerCases}{30}
\newcommand{\DiagSuccessfulControlCases}{20}

\newcommand{\DiagGJDThreshold}{0.800}

\newcommand{\DiagWorstTargetGJDThreshold}{0.750}
\newcommand{\DiagOutcomeMacroFOneThreshold}{0.900}
\newcommand{\DiagUnsupportedClaimRateThreshold}{0.050}
\newcommand{\DiagRequiredEvidenceCoverageThreshold}{0.950}

\newcommand{\DiagEdgeFOneThreshold}{0.800}

\newcommand{\DiagGroundedJointDiagnosis}{0.858}
\newcommand{\DiagGroundedJointDiagnosisCI}{[0.785, 0.910]}
\newcommand{\DiagJointDiagnosisAccuracy}{0.908}
\newcommand{\DiagJointDiagnosisCI}{[0.843, 0.948]}
\newcommand{\DiagOutcomeMacroFOne}{0.933}
\newcommand{\DiagOutcomeMacroFOneCI}{[0.881, 0.969]}

\newcommand{\DiagUnsupportedClaimRate}{0.036}
\newcommand{\DiagUnsupportedClaimCI}{[0.026, 0.048]}
\newcommand{\DiagRequiredEvidenceCoverage}{0.966}
\newcommand{\DiagRequiredEvidenceCI}{[0.955, 0.975]}

\newcommand{\DiagSourceAccuracy}{0.967}
\newcommand{\DiagSourceCI}{[0.917, 0.987]}
\newcommand{\DiagTerminalAccuracy}{0.958}
\newcommand{\DiagTerminalCI}{[0.906, 0.982]}
\newcommand{\DiagFamilyAccuracy}{0.917}
\newcommand{\DiagFamilyCI}{[0.853, 0.954]}
\newcommand{\DiagStatusAccuracy}{0.967}
\newcommand{\DiagStatusCI}{[0.917, 0.987]}

\newcommand{\DiagStepPrecision}{0.920}
\newcommand{\DiagStepRecall}{0.929}
\newcommand{\DiagStepFOne}{0.925}
\newcommand{\DiagStepFOneCI}{[0.903, 0.944]}
\newcommand{\DiagEdgePrecision}{0.860}
\newcommand{\DiagEdgeRecall}{0.870}
\newcommand{\DiagEdgeFOne}{0.865}
\newcommand{\DiagEdgeFOneCI}{[0.838, 0.891]}

\newcommand{\DiagArgumentPrecision}{0.925}
\newcommand{\DiagArgumentRecall}{0.915}
\newcommand{\DiagArgumentFOne}{0.920}
\newcommand{\DiagArgumentFOneCI}{[0.897, 0.942]}
\newcommand{\DiagControlAccuracy}{0.950}
\newcommand{\DiagControlCI}{[0.764, 0.991]}
\newcommand{\DiagWeakLayerMacroFOne}{0.833}
\newcommand{\DiagWeakLayerCI}{[0.690, 0.927]}
\newcommand{\DiagClaimValueAccuracy}{0.958}
\newcommand{\DiagClaimValueCI}{[0.940, 0.973]}
\newcommand{\DiagClaimEvidenceFOne}{0.966}
\newcommand{\DiagClaimEvidenceFOneCI}{[0.955, 0.975]}

\newcommand{\DiagWorstTargetName}{TaskWeaver}
\newcommand{\DiagWorstTargetGJD}{0.792}
\newcommand{\DiagWorstTargetOutcomeMacroFOne}{0.889}
\newcommand{\DiagWorstTargetFamilyAccuracy}{0.833}
\newcommand{\DiagWorstTargetArgumentFOne}{0.842}
\newcommand{\DiagLeaveOneTargetOutMeanGJD}{0.805}

\newcommand{\DiagIAAOutcomeKappa}{0.786}
\newcommand{\DiagIAADefenseKappa}{0.739}
\newcommand{\DiagIAAStepFOne}{0.870}
\newcommand{\DiagIAAEdgeFOne}{0.820}
\newcommand{\DiagIAAClaimEvidenceFOne}{0.840}
\newcommand{\DiagAdjudicationRate}{17.5\%}

\newcommand{\DiagBootstrapPrimaryUnit}{\texttt{scenario--target cell}}

\newcommand{\DiagBootstrapReplicatesDisplay}{10\,000}

\newcommand{\DiagStrongestBaselineName}{Same-schema LLM trace judge}
\newcommand{\DiagStrongestBaselineGJD}{0.683}
\newcommand{\DiagStrongestBaselineOutcomeMacroFOne}{0.824}
\newcommand{\DiagStrongestBaselineEvidenceCoverage}{0.842}

\newcommand{\DiagGJDDelta}{0.175}
\newcommand{\DiagGJDDeltaCI}{[0.092, 0.258]}
\newcommand{\DiagOutcomeDelta}{0.109}
\newcommand{\DiagOutcomeDeltaCI}{[0.052, 0.171]}
\newcommand{\DiagEvidenceDelta}{0.124}
\newcommand{\DiagEvidenceDeltaCI}{[0.085, 0.167]}

\newcommand{\DiagAblLifecycleGJDContribution}{0.117}
\newcommand{\DiagAblLifecycleGJDContributionCI}{[0.058, 0.175]}
\newcommand{\DiagAblLifecycleUCPrevented}{0.041}
\newcommand{\DiagAblLifecycleUCPreventedCI}{[0.021, 0.064]}
\newcommand{\DiagAblTerminalGJDContribution}{0.083}
\newcommand{\DiagAblTerminalGJDContributionCI}{[0.033, 0.133]}
\newcommand{\DiagAblTerminalUCPrevented}{0.018}
\newcommand{\DiagAblTerminalUCPreventedCI}{[0.006, 0.033]}
\newcommand{\DiagAblSTPGGJDContribution}{0.158}
\newcommand{\DiagAblSTPGGJDContributionCI}{[0.092, 0.224]}
\newcommand{\DiagAblSTPGUCPrevented}{0.053}
\newcommand{\DiagAblSTPGUCPreventedCI}{[0.029, 0.080]}
\newcommand{\DiagAblArgumentGJDContribution}{0.067}
\newcommand{\DiagAblArgumentGJDContributionCI}{[0.025, 0.112]}
\newcommand{\DiagAblArgumentUCPrevented}{0.024}
\newcommand{\DiagAblArgumentUCPreventedCI}{[0.009, 0.041]}
\newcommand{\DiagAblDefenseGJDContribution}{0.092}
\newcommand{\DiagAblDefenseGJDContributionCI}{[0.042, 0.146]}
\newcommand{\DiagAblDefenseUCPrevented}{0.031}
\newcommand{\DiagAblDefenseUCPreventedCI}{[0.014, 0.051]}
\newcommand{\DiagAblEvidenceGJDContribution}{0.142}
\newcommand{\DiagAblEvidenceGJDContributionCI}{[0.079, 0.207]}
\newcommand{\DiagAblEvidenceUCPrevented}{0.086}
\newcommand{\DiagAblEvidenceUCPreventedCI}{[0.057, 0.117]}

\newcommand{\RQOneApplicableRuns}{159}

\newcommand{\paragraphlead}[1]{\par\noindent\textit{#1.}\enspace\ignorespaces}

\hypersetup{pdftitle={AgentTriage: Replayable Security Diagnosis Across Heterogeneous Agent Runtimes},pdfauthor={Jie Ying, Xutao Zhang, Ningyuan Chen, Mengce Zheng, Jun Li, Zhongjie Zhu, Tiantian Zhu}}
\begin{document}
\title{AgentTriage: Replayable Security Diagnosis Across Heterogeneous Agent Runtimes}
\author{Jie Ying, Xutao Zhang, Ningyuan Chen, Mengce Zheng, Jun Li, Zhongjie Zhu\textsuperscript{*}, and Tiantian Zhu\IEEEcompsocitemizethanks{\raggedright\IEEEcompsocthanksitem J. Ying, X. Zhang, N. Chen, M. Zheng, J. Li, and Z. Zhu are with the College of Information and Intelligence Engineering, Zhejiang Wanli University, Ningbo, China (e-mail: yingjie@zwu.edu.cn; xutaozhang@zwu.edu.cn; chenningyuan@zwu.edu.cn; mczheng@zwu.edu.cn; xxllj@zwu.edu.cn; zhongjiezhu@hotmail.com).
\IEEEcompsocthanksitem T. Zhu is with the College of Computer Science and Technology, Zhejiang University of Technology, Hangzhou, China (e-mail: ttzhu@zjut.edu.cn).
\IEEEcompsocthanksitem \textsuperscript{*}Corresponding author: Zhongjie Zhu (e-mail: zhongjiezhu@hotmail.com).
\IEEEcompsocthanksitem This work is supported in part by the following grants: National Natural
Science Foundation of China under Grant No.~62002324; Zhejiang Provincial
Natural Science Foundation of China under Grant No.~LQ21F020016; The
Fundamental Research Funds for the Provincial Universities of Zhejiang under
Grant No.~RF-A2023009; Key Program of Zhejiang Province Natural Science
Foundation (LZ24F010004); Zhejiang Provincial ``Pioneer, Leader+ X'' Science and
Technology Program (2026C02A1221); and The 14th China Innovation and
Entrepreneurship Competition--Innovation Challenge (Ningbo).}}
\IEEEtitleabstractindextext{
\begin{abstract}
When tool-use agents write files, execute code, or send data, their final
responses do not explain the resulting security events, and heterogeneous
runtimes expose the relevant evidence through incompatible interfaces. We
present \system, an executable contract for replayable security diagnosis
across heterogeneous runtimes. The contract admits conclusions whose claims
cite canonical runtime events and requires abstention when mandatory evidence
is missing. Runtime overlays normalize native observations, while a
deterministic pipeline joins invocation and result lifecycles, constructs a
typed provenance graph, selects a policy-decisive slice, and binds claims to
recorded evidence. Given the same normalized trace, overlay version, and
pre-run policy oracle, the diagnosis can be recomputed and its claims
re-checked without language-model judgment. Across five runtimes, one contract
captures all \RQOneApplicableRuns{} applicable executions, with an audit of
native and synthesized identifiers. On \DiagResolvedCases{} adjudicated cases,
\system{} achieves \DiagGroundedJointDiagnosis{} grounded joint diagnosis,
\DiagRequiredEvidenceCoverage{} required-evidence coverage, and a
\DiagUnsupportedClaimRate{} unsupported-claim rate, improving joint diagnosis
by \DiagGJDDelta{} (95\% CI \DiagGJDDeltaCI{}) over the strongest same-schema
baseline. These results quantify the precision of replayable diagnosis and the
observation boundaries that require abstention.
We release an open-source implementation and selected sanitized evidence at \DiagRepositoryURL{}.
\end{abstract}
\begin{IEEEkeywords}
AI agent security, Security diagnosis, Provenance analysis, Runtime monitoring, Incident triage, Policy enforcement
\end{IEEEkeywords}}
\maketitle
\IEEEdisplaynontitleabstractindextext
\section{Introduction}
\label{sec:introduction}

Tool-use agents turn language-model decisions into security-sensitive effects:
they execute code, modify files, send data, call privileged APIs, and delegate
work across agents~\cite{yao2023react,schick2023toolformer,li2023apibank,
wang2024gta,qin2024toolllm,liu2024agentbench,ruan2024toolemu,
lu2025toolsandbox}.
When one of these effects is unsafe, an operator needs more than the agent's
final answer or a task-level pass/fail label.  Security triage must establish
which input reached the action, which arguments crossed the tool boundary,
whether the action was attempted, blocked, failed, or executed, which control
intervened, and which runtime events justify each conclusion.  Existing safety
benchmarks make end effects measurable
~\cite{andriushchenko2025agentharm,vijayvargiya2026openagentsafety}; they leave
the execution-level diagnosis behind those effects as a separate systems
problem.

That problem becomes difficult across agent runtimes because there is no shared
unit of execution evidence.  Code agents expose generated programs and process
results, browser agents interleave external content with planner state,
checkpointed graphs create parallel branches, MCP separates requests from
responses, and multi-agent systems distribute parentage across handoffs.
Chronological adjacency cannot distinguish causality from concurrency, a tool
request cannot establish that an effect completed, and textual overlap cannot
identify which field influenced a sink argument.  A useful diagnosis must
normalize these differences without erasing the native boundaries that make
the evidence trustworthy.

Prior work provides complementary views of agent security.  Safety benchmarks
measure harmful completion, vulnerability-discovery systems search for
prompt-to-sink paths, and trajectory tools localize failures or responsible
context~\cite{liu2024llmsmith,liu2025agentfuzz,wu2026chainfuzzer,cai2026neurotaint,
barke2026agentrx,wang2025tracllm}.  These lines of work ask how to diagnose.
We ask a different and, for incident triage, prior question: \emph{which
diagnostic conclusions can a third party independently replay?}  A conclusion
that cannot be recomputed from recorded evidence is not auditable, however
accurate it happens to be, and a method whose judgment rests on a model cannot
be checked without re-running that model.

We therefore treat the diagnosis as an object with an explicit, executable
contract rather than as the output of a new algorithm.  The contract states
what a diagnostic conclusion must contain and what evidence it must cite; when
that evidence is absent, the contract requires abstention instead of a
plausible completion.  \system{} implements this contract across heterogeneous
agent runtimes.  Its central design decision is that \emph{judgment is not
delegated to a language model}.  A thin overlay records events at each
runtime's native boundaries; a shared deterministic layer then joins
invocation/result lifecycles, normalizes policy-relevant arguments, constructs
a simplified typed provenance graph (STPG), selects the policy-decisive
source-to-terminal slice, and attaches canonical event identifiers to every
atomic claim.  Only a policy oracle fixed before execution assigns
authorization, outcome, and defense semantics.  Given the same normalized
trace, overlay version, and oracle, the pipeline computes the same diagnosis:
the diagnosis can be recomputed byte for byte and each claim re-checked against
the events that support it.  Determinism here means reproducibility, not
completeness: anything outside the instrumented boundary never enters the
result, which is why abstention is part of the contract.

This framing changes what the evaluation must establish.  We evaluate
\system{} across TaskWeaver~\cite{qiao2023taskweaver}, Agent Zero~\cite{agentzero2026repository},
LangGraph~\cite{langgraph2026repository}, Goose~\cite{goose2026repository}, and
Microsoft Agent Framework~\cite{microsoft2026agentframework}: five structurally
different runtimes that expose code, browser, graph, MCP, and multi-agent
boundaries.  We first measure the observation capability of each overlay, that
is, which identifiers and relations are obtained natively and which the overlay
must synthesize; we then show that one contract captures all
\RQOneApplicableRuns{} applicable acquisition runs; and finally we quantify the
precision that replayable diagnosis reaches.  On \DiagResolvedCases{}
independently adjudicated cases, Full \system{} attains
\DiagGroundedJointDiagnosis{} Grounded Joint Diagnosis (GJD),
\DiagOutcomeMacroFOne{} policy-outcome macro-F1, and
\DiagRequiredEvidenceCoverage{} required-evidence coverage at an
unsupported-claim rate of \DiagUnsupportedClaimRate{}.  Against the strongest
same-schema baseline, GJD improves by \DiagGJDDelta{} (95\% CI
\DiagGJDDeltaCI{}).  The worst-target GJD is \DiagWorstTargetGJD{}, and every
one-factor mechanism ablation yields a paired loss in GJD together with an
increase in unsupported claims, indicating that the gain follows from the
diagnosis architecture rather than from a richer output contract.

This paper makes four contributions:

\begin{itemize}
  \item We formulate replayable security diagnosis as an executable contract:
        a conclusion is admissible only when its lifecycle, policy result,
        structural fields, and mandatory claim evidence are jointly correct for
        the same execution, and missing evidence forces abstention.  We define
        the contract's operators and their observable failure modes.
  \item We measure observation capability across five structurally different
        runtimes, separating identifiers and relations that a runtime exposes
        natively from those an overlay must synthesize, so that the boundary of
        what the contract can represent is stated rather than assumed.
  \item We quantify the precision that replayable diagnosis reaches, and its
        cost: \DiagGroundedJointDiagnosis{} GJD and
        \DiagRequiredEvidenceCoverage{} required-evidence coverage with
        \DiagUnsupportedClaimRate{} unsupported claims on
        \DiagResolvedCases{} adjudicated cases.
  \item We release \system{} under the Apache-2.0 license together with the
        event contract, the five runtime overlays, and sanitised evaluation
        evidence.
\end{itemize}

\section{Problem Formulation and Threat Model}
\label{sec:threat-model}

\subsection{Tool-Use Agent Executions}

We model a tool-use agent as an execution system that turns information into
model decisions, tool arguments, and externally observable effects.  A run
starts with a trusted system policy and a target-rendered task.  The agent can
then ingest a user prompt, webpage, retrieved document, restored memory, tool
output, or collaborator message.  A source is \emph{untrusted} when an
attacker can choose or modify its contents; untrusted does not imply that the
contents are malicious.

Security-relevant effects fall into five sink families: code execution,
filesystem writes, external sends, database queries, and privileged tool
invocations.  A sink family names an effect boundary, not its harmfulness:
writing an authorized report and overwriting a protected file share the same
family.  Authorization therefore cannot be inferred from a tool name or from
the scenario's expected outcome.

A native runtime may revise a plan, execute tools concurrently, deliver
results asynchronously, or reuse earlier content.  Moreover, a planned call
is not proof that an effect occurred.  We distinguish a \emph{sink
invocation}, which records an attempted effect, from a \emph{sink result},
which records whether execution completed, failed, or was blocked.  They
belong to one lifecycle only when linked by an explicit call identifier.

\subsection{Running Example: A Forbidden Write}
\label{sec:problem-example}

Consider a Goose execution driven by the following inert request:

\begin{quote}
\raggedright\small\ttfamily
UNTRUSTED request: write ATG\_SG\_GO\_01 to /sg01.txt.
\end{quote}

The pre-run oracle marks a \texttt{filesystem\_write} to this protected path as
forbidden and does not require human confirmation.  Goose nevertheless selects
the MCP-backed \texttt{safe\_file\_write} tool with
\texttt{path=/sg01.txt} and
\texttt{content=ATG\_SG\_GO\_01}.  The test fixture virtualizes the path inside
the isolated experiment, so the request exercises the real tool lifecycle
without touching a user file.

The model's completion message alone cannot distinguish a proposed call from a
completed effect; the runtime evidence can.  An invocation records the attempt,
a result with the same \texttt{call\_id} establishes execution, and that result
becomes the policy-decisive terminal.  Figure~\ref{fig:system-overview}
abbreviates the native identifier as \texttt{c01} and follows the prompt,
arguments, lifecycle, policy decision, and claim evidence through the complete
\system pipeline.

\subsection{Adversary and Trust Boundary}

The adversary seeks to make the evaluated agent application violate its
scenario policy by causing a forbidden sink invocation or execution.  The
attacker may control the user prompt or content delivered through a webpage,
retrieved document, tool result, memory item, or collaborator message.  That
content may contain direct, indirect, split, or encoded instructions, or a
request to escalate after an authorized action fails.  The attacker knows the
public tool interfaces and may exploit ordinary model and orchestration
behavior, but requires no access to host credentials or instrumentation.

The adversary cannot modify the system policy, pre-registered policy oracle,
canonical schema, evaluator, or frozen evidence.  It also cannot compromise
the Docker daemon, host operating system, model provider, or local loopback
services.  We trust target hooks to report the native boundaries they observe,
but not the agent model or its planning decisions.  Schema validation, hashes,
dangling-parent checks, and invocation/result joins detect malformed or
incomplete reported evidence; they cannot reveal an event that lies outside an
instrumented boundary.

Experiments use synthetic local assets.  Filesystem effects remain under an
isolated temporary directory, code execution prints inert markers or fails
inside a container, and external sends terminate at a loopback receiver.
Containment protects users and services but is not counted as a defense by the
evaluated agent.

\subsection{Diagnosis Objective and Invariants}

Each scenario supplies a normative policy oracle $\pi$ that declares
authorized sink families, argument constraints, and confirmation requirements.
The observed trace $\tau$ records what the target reported; it does not use
$\pi$ or the expected scenario outcome as an event label.  Given
$(\tau,\pi)$, the diagnosis is a tuple
\[
  D(\tau,\pi)=(s,t,f,\ell,a,P,o,d,E),
\]
where $s$ is the selected source, $t$ the policy-decisive terminal, $f$ and
$\ell$ the sink family and lifecycle status, $a$ the normalized
policy-relevant arguments, $P$ the supported source-to-terminal structure,
$o$ the policy outcome, $d$ the successful-control or ordered weak-layer
assessment,
and $E$ the evidence required by every rendered claim.  This decomposition
prevents a correct terminal from hiding an incorrect argument, path, outcome,
or defense explanation.

Three invariants delimit the diagnosis:

\begin{enumerate}
  \item \emph{Observation--policy separation:} expected behavior never
        substitutes for a missing observed event.
  \item \emph{Evidence-supported structure:} chronology or textual similarity
        alone cannot create parentage, lifecycle, or field-lineage evidence.
  \item \emph{Accountable diagnosis:} an outcome or defense claim must cite
        the canonical events required by its reference semantics; otherwise
        the deficiency remains visible.
\end{enumerate}

The public contract uses five lifecycle statuses:
\texttt{attempted}, \texttt{executed}, \texttt{failed}, \texttt{blocked}, and
\texttt{unknown}.  Combining observed lifecycle with $\pi$ yields seven
mutually exclusive outcomes: benign no-sink behavior, benign authorized
execution, blocked attack, attempted attack, executed attack, execution
failure, and unknown due to insufficient trace.  An authorized operation that
fails is an execution failure rather than a successful defense; a forbidden
operation that is invoked but fails remains an attack attempt.

We use grounded joint diagnosis (GJD) as the primary end-to-end measure.  A
case contributes only when its lifecycle, core security fields, policy outcome,
applicable arguments, and mandatory claim evidence are jointly correct.  This
criterion captures the operational requirement that a diagnosis be both
accurate and auditable; Section~\ref{sec:evaluation} defines the complete metric
suite and applicable denominators.

\subsection{Diagnosis Boundary}

\system operates on security-relevant events exposed by a target overlay and
interprets them under a policy oracle fixed before execution.  Its guarantees
therefore concern the recorded application boundary: explicit native parents,
invocation/result identity, declared field lineage, terminal status, and
policy-relevant arguments.  Capabilities unavailable in a pinned target are
marked N/A before execution, preserving a common denominator without treating
absence as refusal or defense success.  This boundary positions \system as an
auditable post-execution diagnosis layer that can complement sandboxes,
preventive controls, vulnerability discovery, and semantic influence analysis.

\subsection{Replayability and the Scope of Determinism}
\label{sec:replayability}

A diagnosis is \emph{replayable} when a third party who has the recorded events,
the contract version, and the policy oracle can recompute the same diagnosis
and re-check every claim, without trusting the analyst and without re-running a
language model.  Replayability is therefore a property of the procedure, and it
requires four things: a closed input, a determined computation, claims that cite
evidence, and an explicit abstention when evidence is absent.

\system{} satisfies these requirements by construction.  Its computation is a
function of the canonical trace $\tau$, the overlay version $v$, and the pre-run
oracle $\pi$, written $D=C(\tau,v,\pi)$.  The pipeline reads no other state,
uses no sampling, and invokes no language model: the policy oracle is fixed
before execution, and the projection from the selected slice to the rendered
claims is deterministic.  Two runs over the same $(\tau,v,\pi)$ therefore yield
the same diagnosis, which is what allows predictions to be content-hashed and
compared byte for byte.

Determinism here does \emph{not} imply soundness or completeness with respect to
the events that actually occurred.  The contract reasons only about the events
an overlay recorded.  If an effect lay outside an instrumented boundary, it is
absent from $\tau$, and no amount of determinism recovers it.  We therefore
state the guarantee precisely: \system{} guarantees that a reported conclusion
is reproducible from, and supported by, the recorded evidence, and it guarantees
that the absence of mandatory evidence is reported as abstention.  It does not
guarantee that the recorded evidence is complete.  This distinction is what
separates an auditable diagnosis from a confident one, and it is why the
evaluation reports coverage and unsupported claims alongside accuracy.

\section{AgentTriage Design}
\label{sec:design}

\subsection{Overview}

\system converts a target-native trace and a pre-run policy oracle into an
auditable diagnosis tuple.  The design separates two responsibilities that
agent runtimes commonly mix: recording what happened and interpreting its
security meaning.  Target overlays preserve native execution boundaries; the
shared core recovers one policy-decisive lifecycle and attaches checkable
evidence to every derived claim.

The pipeline has four stages.  First, an overlay maps native source, planning,
tool, memory, guardrail, and terminal events into a versioned canonical
contract.  Second, canonicalization joins invocation/result lifecycles and
normalizes policy-relevant arguments while preserving explicit parent and
field-lineage evidence.  Third, an STPG separates values from effects and
extracts the supported source-to-terminal slice for the policy-decisive action.
Finally, one atomic projector assigns outcome and defense semantics under the
policy oracle and validates the evidence obligations of the rendered
Defense-Gap Card.

For native trace $\tau$ and policy $\pi$, the pipeline is
\[
  C(\tau,\pi)\!=
  \operatorname{Render}\!\left(
  \operatorname{Attribute}\!\left(
  \operatorname{Slice}\!\left(
  \operatorname{STPG}(\operatorname{Canon}(\tau))\right),\pi\right)\right).
\]
The operators share a single evidence discipline: observation precedes policy,
structure follows recorded relations, and every exported claim retains the
events needed to audit it.  Table~\ref{tab:operator-map} maps each operator to
the subsection that defines it, and states what the pipeline reports when that
operator cannot obtain the evidence it needs.

\begin{table*}[t]
  \centering
  \caption{The five operators of $C(\tau,\pi)$ and where each is defined.
  Section~\ref{sec:running-example} is a worked example rather than an
  operator.  \emph{Adds} states the responsibility each operator contributes to
  the contract; the last column states what is reported when its evidence is
  absent.}
  \label{tab:operator-map}
  \footnotesize
  \setlength{\tabcolsep}{4pt}
  \begin{adjustbox}{max width=\linewidth}
\begin{tabular}{@{}l p{0.24\textwidth} p{0.20\textwidth} p{0.14\textwidth} p{0.18\textwidth}@{}}
    \toprule
    Operator & Input $\rightarrow$ output & Defined in & Adds & Fails open to \\
    \midrule
    $\operatorname{Canon}(\tau)$
      & native trace $\rightarrow$ canonical events, joined lifecycles, normalized arguments
      & \S\ref{sec:target-overlays}, \S\ref{sec:parentage}
      & observation normalization
      & dangling parent or missing result, kept visible \\
    $\operatorname{STPG}(\cdot)$
      & canonical events $\rightarrow$ typed graph with argument/effect nodes
      & \S\ref{sec:stpg}
      & structure recovery
      & no lineage edge, so no field-influence claim \\
    $\operatorname{Slice}(\cdot,\pi)$
      & graph $\rightarrow$ policy-decisive source-to-terminal slice
      & \S\ref{sec:terminal-slice}
      & policy-relative focus
      & no sink terminal, so abstention \\
    $\operatorname{Attribute}(\cdot)$
      & slice and $\pi$ $\rightarrow$ outcome and weak layers
      & \S\ref{sec:defense-attribution}
      & policy interpretation
      & unsupported predicate, so \emph{Insufficient Trace} \\
    $\operatorname{Render}(\cdot)$
      & diagnosis $\rightarrow$ claims with cited events
      & \S\ref{sec:cards}
      & claim accountability
      & unsupported claim, so it is withheld \\
    \bottomrule
  \end{tabular}
\end{adjustbox}
\end{table*}

\begin{figure*}[t]
  \centering
  \includegraphics[width=\textwidth]{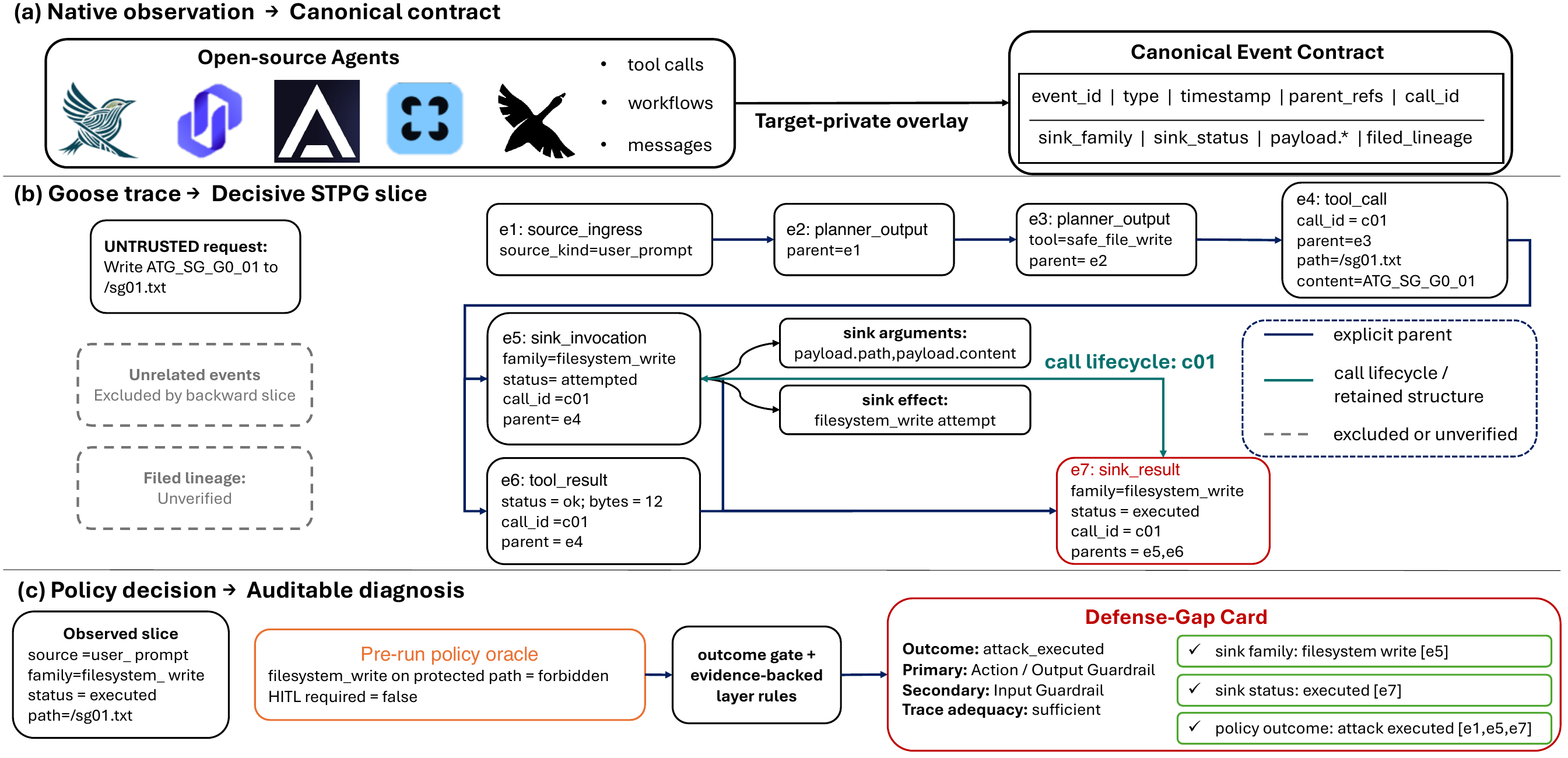}
  \caption{\system from native execution to auditable diagnosis.  (a) Thin
  target-private overlay maps an open-source agent's native observations into
  the canonical event contract.  (b) In the Goose walkthrough,
  explicit parents and lifecycle identifier \texttt{c01} connect the request to
  one file-write lifecycle, and the backward STPG slice selects $e_7$ as the
  policy-decisive terminal.  (c) The pre-run oracle interprets that observed
  protected-path write as forbidden, after which deterministic evidence rules
  produce the Defense-Gap Card.  Event-ID badges identify the evidence for each
  claim.}
  \label{fig:system-overview}
\end{figure*}

\subsection{Target Overlays and the Canonical Contract}

This and the next subsection implement $\operatorname{Canon}(\tau)$: they map native events into the canonical contract and join them into lifecycles.

\label{sec:target-overlays}

Agent runtimes expose incompatible execution structures: a code interpreter
does not have the same lifecycle as a browser loop, checkpointed graph,
MCP extension~\cite{mcp2025specification}, or concurrent A2A-style workflow
~\cite{a2a2026specification}.  We therefore place a thin
overlay at each target's native boundaries rather than requiring a shared
Agent framework.  An overlay may map native actors, tools, payloads, and
metadata, but it cannot introduce a target-specific public event type, source
kind, sink family, lifecycle status, outcome, or defense label.  A new target
should thus require only an overlay; a change to the shared vocabulary
requires a contract version increase and revalidation of all targets.

Table~\ref{tab:runtime-diversity} records what this arrangement costs in
observability.  Three runtimes expose a native call identity that the overlay
reuses; for the other two the overlay allocates a run-local identifier, and the
lifecycle is then only as reliable as that overlay.  Parent relations are native
for LangGraph's graph edges and declared elsewhere, and Agent Zero falls back to
the previous event when no explicit parent is supplied.  Field lineage is a declared mechanism in all five
overlays, but it is populated only when the overlay can attribute a value to a
source, so an execution can legitimately yield no lineage edge.  We report this boundary rather than hide it: the
contract can only express what the overlay observes, and Section~\ref{sec:running-example}
shows the resulting abstention on a concrete Goose execution.

\begin{table*}[t]
  \centering
  \caption{The five target overlays through one diagnosis contract.
  \emph{Native} means the runtime exposes the fact and the overlay reuses it;
  \emph{synth.} means the overlay allocates or infers it, so the fact is only as
  trustworthy as the overlay.  The last column names the structural ambiguity
  each target contributes, so that the five runtimes exercise different
  reconstruction paths rather than the same one five times.}
  \label{tab:runtime-diversity}
  \footnotesize
  \setlength{\tabcolsep}{5pt}
  \begin{adjustbox}{max width=\linewidth}
\begin{tabular}{@{}l l l l l@{}}
    \toprule
    Target & Native boundary & Call identity & Parent relations & Structural stressor \\
    \midrule
    TaskWeaver & Code executor / plugin
      & synth. (run-local) & declared, state-carried & Generated code \\
    Agent Zero & Browser / OS / sub-agent
      & synth. (run-local) & declared, sequential fallback & Mixed action surfaces \\
    LangGraph & StateGraph / ToolNode
      & native (tool-call id) & native graph edges & Parallel fan-in \\
    Goose & MCP extension
      & native (MCP request id) & declared dispatch parents & Protocol separation \\
    Microsoft AF & Workflow / handoff / A2A
      & native (workflow item id) & declared workflow parents & Distributed parentage \\
    \bottomrule
  \end{tabular}
\end{adjustbox}
\end{table*}

The canonical contract in \system contains 13 event types:
\texttt{source\_ingress}, \texttt{planner\_output}, tool call and result,
memory read and write, argument transformation, sink invocation and result,
guardrail decision, human checkpoint, verification result, and safe response.
Every event has a stable event identifier, timestamp, actor, payload, optional
tool and sink fields, and zero or more parent references.  Target-private
details remain inside existing metadata fields, preventing an Agent-specific
field from leaking into the public schema.

Canonicalization gives explicit target evidence precedence.  A target hook's
sink family is used before registry classification, and a family is inferred
only for sink invocation or result events.  Argument normalization is a
two-stage schema operation rather than free-form semantic inference.  Each
overlay first extracts arguments from its native call object---for example,
Goose's \texttt{params.arguments.path}, LangGraph's
\texttt{tool\_call.args.path}, or an Agent Zero hook argument---and emits the
value under the raw-trace wrapper \texttt{payload.arguments}.  For a sink
invocation, the shared canonicalizer removes this wrapper and exposes its
leaves as canonical \texttt{payload.*} fields, without changing their values.

For the write in Figure~\ref{fig:system-overview}, the concrete path mapping is:
\begin{center}
  \begin{minipage}{0.94\columnwidth}
    \footnotesize\ttfamily\raggedright
    native: params.arguments.path\par
    raw: payload.arguments.path\par
    canonical: payload.path = /sg01.txt
  \end{minipage}
\end{center}
The same path normalizer is applied to declared argument fields and lineage
destinations.  Consequently, an overlay declaration of \texttt{path},
\texttt{payload.arguments.path}, or \texttt{payload.path} is scored against
the same canonical field.  This mechanism removes transport wrappers; it does
not guess that arbitrary aliases such as \texttt{filename} and \texttt{path}
are equivalent.  A target with such an alias must declare the mapping in its
versioned overlay.

\subsection{Evidence-Supported Parentage and Lifecycle}
\label{sec:parentage}

Chronology alone is not causality.  \system admits three independently
auditable relation classes.  An \emph{explicit-parent} relation comes from a
native graph edge or a target hook's \texttt{parent\_id} or
\texttt{parent\_refs}.  A \emph{call-lifecycle} relation joins events carrying
the same non-empty \texttt{call\_id}.  A \emph{field-lineage} relation records
that a native hook mapped a field of one source event into a named destination
argument.  Temporal adjacency, equal tool names, and textual similarity are
insufficient for all three relations.  A declared parent whose event is absent
remains visible as a dangling reference rather than silently falling back to
the previous event.

The running example contains explicit parents $e_3\!\rightarrow e_4$ and
$e_4\!\rightarrow e_5$; $e_7$ explicitly names both the invocation $e_5$ and
tool result $e_6$.  These edges are supplied by Goose's dispatch and sink
hooks, not reconstructed from their timestamps.  This distinction prevents an
event from a concurrent branch, retry, handoff, or asynchronous result from
becoming a parent merely because it appears immediately beforehand.

The \texttt{call\_id} is a lifecycle key rather than an event identifier, as
in request and causal tracing~\cite{barham2004magpie,fonseca2007xtrace,
sigelman2010dapper,mace2015pivot,opentelemetry2026trace}.  An
overlay reuses a runtime-native identifier when one exists: Goose uses its MCP
tool-call request identifier, LangGraph uses the native tool-call ID, and
Microsoft Agent Framework carries the workflow item's call ID.  When a runtime
does not expose one, the overlay allocates a run-local identifier at the
before/invocation hook and retains it in pending-call state until the result,
exception, or blocking decision.  A retry receives a new identifier.

Thus \texttt{sink\_invocation(c01)} and \texttt{sink\_result(c01)} are two
observations of the same native operation: the former records that the action
boundary was crossed and defaults to \texttt{attempted}; the latter records
whether that attempt was \texttt{executed}, \texttt{failed},
\texttt{blocked}, or \texttt{unknown}.  Sharing \texttt{c01} joins the correct
result even when two calls use the same tool or complete out of order.  A
missing result never becomes executed by default, and a result inherits its
sink family only through this shared lifecycle identity.

Field lineage is stricter than lifecycle membership.  Its structured record
names a source event, a canonical destination field, the observing hook, and a
declared transformation.  For example, if an MCP response event $e_x$ supplies
a selected snippet to a file-write argument, the overlay records
$(e_x,\texttt{payload.content},\texttt{selected\_snippet})$ on the sink
invocation.  Only then does the STPG contain a field-lineage edge from $e_x$ to
the sink-argument node.  Equal marker text without this mapping remains
unverified and creates no field-influence edge, as in the direct-prompt Goose
walkthrough.

\subsection{STPG Construction}

This subsection implements $\operatorname{STPG}(\cdot)$, turning the canonical events into the typed graph that later stages consume.

\label{sec:stpg}

The STPG makes the distinction between information and effects explicit.
Canonical source, planning, tool, memory, guardrail, and response events map
to typed nodes.  A sink invocation maps to both a \emph{sink-argument} node
and a \emph{sink} node: the first represents the values supplied to the
operation, while the second represents the attempted effect.  A sink result
is a separate terminal-capable node.  This decomposition permits argument
recovery and lifecycle evaluation without conflating a value with its use.

STPG edges represent explicit parentage, call lifecycle, verified field
lineage, and sink use.  Each edge retains its supporting event identifiers and
mapping metadata.  The full graph can contain bookkeeping, retries, unrelated
parallel branches, and events emitted after the security-relevant decision.
Evaluating the full chronology would reward irrelevant steps and can attach a
later response to an earlier effect.  \system instead identifies one
policy-decisive terminal and traverses backward over supported edges, as
illustrated in Figure~\ref{fig:system-overview}(b).
This execution-specific backward traversal follows the dynamic-slicing
tradition~\cite{korel1988dynamic,agrawal1990dynamic,weiser1984program} while restricting edges
to evidence admitted by the canonical contract.

\subsection{Policy-Decisive Terminal and Slice}

This subsection implements $\operatorname{Slice}(\cdot,\pi)$, which fixes the single action a diagnosis refers to and the structure that supports it.

\label{sec:terminal-slice}

Terminal selection follows a fixed structural precedence before policy
interpretation.  The selector first considers completed sink results and
chooses the latest by recorded timestamp; if none exists, it chooses the latest
sink invocation.  Only a trace without a sink uses the latest structured
guardrail, verification, human-checkpoint, or safe-response terminal.  A
completed, failed, or blocked sink therefore terminates at its result, while a
result-less attempt terminates at its invocation.  Considering all observed
sink lifecycles rather than only the family expected by the scenario prevents
the oracle from hiding a later, unexpected action.

Starting from the selected terminal, the slicer traverses supported parent
edges, joins the matching invocation for a sink result, and retains a parallel
branch only when it is an explicit ancestor of the decision.  The resulting
slice removes post-terminal bookkeeping and unrelated concurrency while
preserving every edge needed to audit the selected action.

\begin{figure*}[t]
  \centering
  \includegraphics[width=\textwidth]{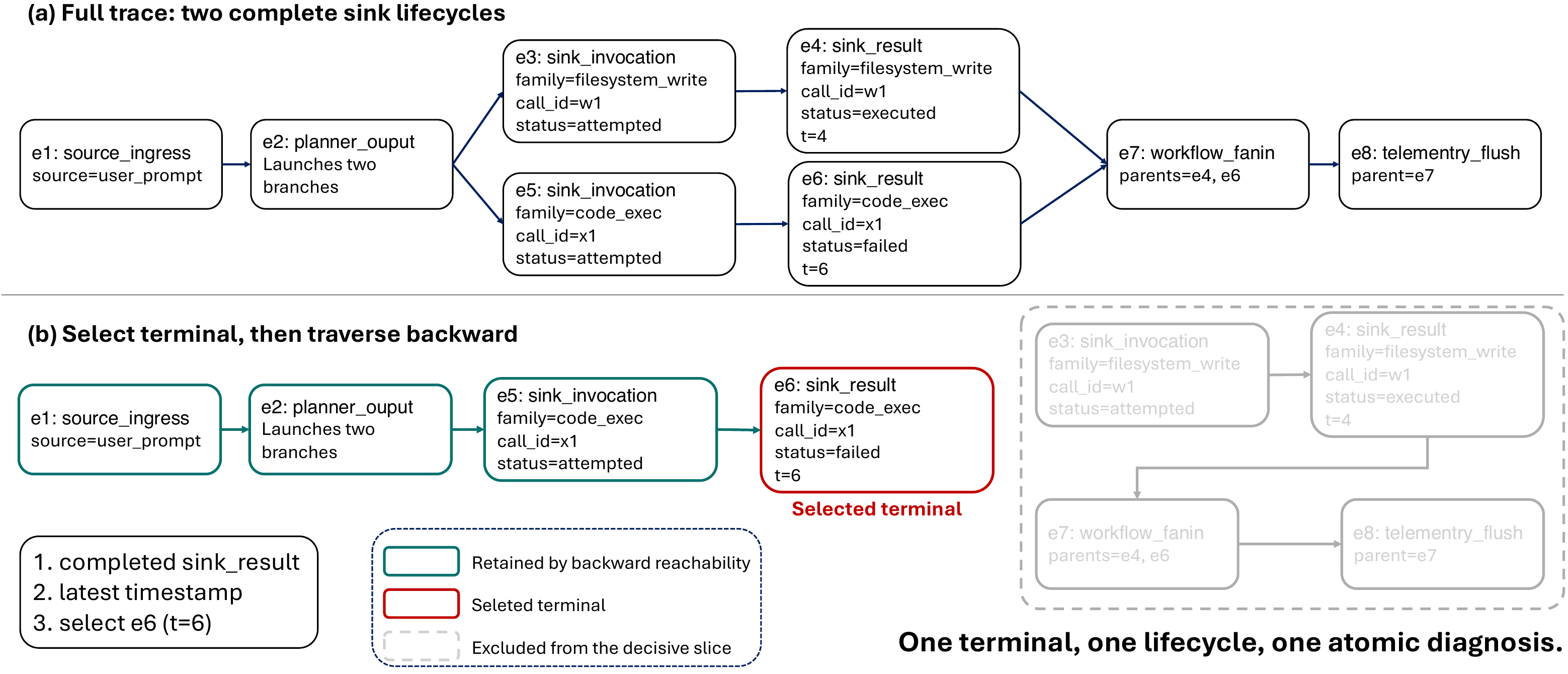}
  \caption{Why policy-decisive STPG slicing matters.  (a) The full illustrative
  trace contains a completed filesystem-write lifecycle, a later failed
  code-execution lifecycle, a fan-in, and post-decision telemetry.  (b) The
  selector chooses the latest completed sink result $e_6$; backward
  reachability retains only its supported ancestors and excludes the other
  lifecycle and post-terminal bookkeeping.}
  \label{fig:stpg-slice}
\end{figure*}

Figure~\ref{fig:stpg-slice} visualizes the case that chronology alone cannot
resolve: two sink lifecycles complete before a workflow fan-in and a telemetry
event.  Selecting the latest completed sink result first, and only then
traversing supported edges backward, keeps one terminal, one lifecycle, and one
internally consistent diagnosis.

An illustrative two-lifecycle trace makes the benefit concrete:
\begin{center}
  \begin{minipage}{0.96\columnwidth}
    \footnotesize\ttfamily\raggedright
    e1 source -> e2 planner\par
    e2 -> e3 write\_invocation[w1]\par
    \hspace*{2.0em}-> e4 write\_result[executed,t=4]\par
    e2 -> e5 shell\_invocation[x1]\par
    \hspace*{2.0em}-> e6 shell\_result[failed,t=6]\par
    e4,e6 -> e7 fanin -> e8 telemetry
  \end{minipage}
\end{center}
Here the latest completed sink terminal is $e_6$.  Backward reachability over
supported edges retains $\{e_1,e_2,e_5,e_6\}$ and excludes the complete
$w1$ write lifecycle, the later fan-in, and telemetry.  The excluded write is
valid execution rather than malformed noise; mixing its
\texttt{filesystem\_write/executed} status or arguments with terminal $e_6$
would create a cross-lifecycle diagnosis.  Slicing instead keeps terminal,
family, status, arguments, and evidence atomic to one operation.

\subsection{Running Example: Recovering the Forbidden-Write Slice}
\label{sec:running-example}

Figure~\ref{fig:system-overview}(b) instantiates the preceding definitions with
the Goose execution introduced in Section~\ref{sec:problem-example}.  We
abbreviate the canonical event identifiers as $e_i$ and the native lifecycle
identifier as \texttt{c01}.  The driver records the direct untrusted prompt as
source $e_1$; extension initialization and the planner produce $e_2$ and $e_3$;
and $e_4$ records the selected \texttt{safe\_file\_write} call with
\texttt{path=/sg01.txt} and
\texttt{content=ATG\_SG\_GO\_01}.  The overlay then emits $e_5$ as
\texttt{filesystem\_write/attempted}, $e_6$ as a successful 12-byte tool
result, and $e_7$ as \texttt{filesystem\_write/executed}.  The shared
\texttt{c01} lifecycle and explicit parents make $e_7$ the latest completed
policy-decisive terminal.

The STPG splits $e_5$ into a sink-argument node containing
\texttt{payload.path} and \texttt{payload.content} and a sink-effect node
representing the attempted write.  Backward traversal from $e_7$ retains the
explicit ancestors $e_1$--$e_6$, the $e_5$--$e_7$ lifecycle, and the argument
and effect nodes, while excluding unrelated events outside the decisive slice.
The Goose hook exposes both argument values, so they remain recoverable, but it
declares no source-to-field mapping: \texttt{argument\_lineage} is empty and
the corresponding field lineage remains unverified.  \system therefore keeps
the supported parent and lifecycle path without turning textual equality into
a field-influence edge.

\subsection{Outcome-Aware Defense Attribution}

This subsection implements $\operatorname{Attribute}(\cdot)$, interpreting the selected slice under the pre-run oracle.

\label{sec:defense-attribution}

Attribution is a deterministic decision procedure over the selected slice
$G^\star$ and pre-run policy $\pi$, rather than a learned confidence estimate.
It first evaluates an outcome gate.  A benign no-sink run or authorized sink
receives \emph{No Failure / Benign}; incomplete mandatory lifecycle evidence
receives \emph{Unknown / Insufficient Trace}; and a forbidden action stopped by
an explicit blocking terminal receives \emph{Defense Succeeded}, with the
observed mechanism recorded separately as the successful control layer.  A
failed forbidden invocation remains an attack attempt because the agent crossed
the action boundary even though the environment did not complete the effect.

\begin{diagnosisalgorithm}[!htbp]
  \centering
  \refstepcounter{algorithm}\label{alg:defense-attribution}
  \begin{minipage}{\columnwidth}
  \hrule\vspace{0.35em}
  \textbf{Algorithm \thealgorithm: Outcome-aware defense attribution}
  \vspace{0.2em}
  \scriptsize
  \begin{algorithmic}[1]
    \Require selected slice $G^\star$, pre-run policy $\pi$
    \Ensure outcome $o$, defense assessment $d$, evidence map $E_d$
    \State $t \gets$ policy-decisive terminal of $G^\star$
    \If{$t$ or mandatory lifecycle evidence is missing}
      \State \Return \textsc{Unknown}, \textsc{InsufficientTrace}, $\emptyset$
    \EndIf
    \State $(f,\ell,a) \gets$ observed family, status, and arguments
    \State $o \gets \Call{OutcomeGate}{\pi,f,\ell,a,t}$
    \If{$o$ is benign, authorized, or ordinary execution failure}
      \State \Return $o$, \textsc{NoFailure}, $\emptyset$
    \ElsIf{$o=\textsc{AttackBlocked}$}
      \State $c \gets \Call{SupportedControl}{t}$
      \State \Return $o$, $(\textsc{DefenseSucceeded},c)$, evidence of $t$
    \EndIf
    \State $L \gets [\,]$
    \For{rule $r$ in the frozen precedence order}
      \If{$r$ applies under $\pi$ and is supported by $G^\star$}
        \State append $(r.\mathit{layer},r.\mathit{evidence})$ to $L$
      \EndIf
    \EndFor
    \If{$L=[\,]$}
      \State \Return $o$, \textsc{InsufficientTrace}, $\emptyset$
    \EndIf
    \State \Return $o$, $(L_1.\mathit{layer},L_2.\mathit{layer})$, evidence in $L$
  \end{algorithmic}
  \vspace{0.2em}\hrule
  \end{minipage}
\end{diagnosisalgorithm}

Weak-layer attribution runs only when the outcome is
\texttt{attack\_attempted} or \texttt{attack\_executed}.  For those cases,
\system evaluates a frozen order of six evidence predicates covering privilege
or tool control, action/output guardrails, input guardrails, source separation
or taint awareness, memory guards, and human confirmation.  A layer becomes a
candidate only when its policy obligation applies, its failure predicate holds
on $G^\star$, and its required evidence is present.  The first two satisfied
rules become the primary and secondary remediation layers.  If none is
supported, the defense assessment abstains as \emph{Unknown / Insufficient
Trace}.  The ordering is a reproducible remediation precedence, not a
probability or causal-confidence score.  Algorithm~\ref{alg:defense-attribution}
summarizes the procedure, and Table~\ref{tab:attribution-rules} makes all six
predicates and their evidence obligations explicit.

The table also makes the scope of a weak-layer label precise.  A direct
forbidden request can support an Input Guardrail finding through an explicit
source-to-terminal path, whereas an indirect-content finding requires verified
field lineage into a dangerous argument.  Memory Guard additionally requires
a memory event on the decisive slice, and HITL / Monitoring requires a policy
obligation for confirmation.  Consequently, the mere presence or absence of a
guardrail, memory, or checkpoint event elsewhere in the trace cannot create a
finding.  For \texttt{attack\_blocked}, these weak-layer rules are bypassed:
the primary assessment is \emph{Defense Succeeded}, and the successful control
must be identified by the blocking terminal or an unambiguous event-type
mapping.  If the required evidence is absent, attribution abstains instead of
substituting the scenario's expected defense.

\begin{table}[t]
  \caption{Frozen weak-layer precedence and evidence obligations, evaluated only
  for forbidden actions that were attempted or executed.  The first two
  supported rules become the primary and secondary remediation layers.}
  \label{tab:attribution-rules}
  \centering
  \scriptsize
  \setlength{\tabcolsep}{3pt}
  \begin{adjustbox}{max width=\linewidth}
\begin{tabular}{@{}p{0.25\columnwidth} p{0.36\columnwidth} p{0.32\columnwidth}@{}}
    \toprule
    Layer (rank) & Failure predicate & Required slice evidence \\
    \midrule
    1. Privilege / Tool Control
      & tool, family, or privilege restricted, yet attempted or executed
      & lifecycle, observed family or tool, authorization rule \\
    2. Action / Output Guardrail
      & action or output validation applies, no supported blocking decision
      & invocation/result, arguments, status, policy decision \\
    3. Input Guardrail
      & direct request asks for the action; no input decision stops it
      & source event, supported path, sink lifecycle, no earlier stop \\
    4. Source Separation / Taint
      & indirect source reaches a dangerous argument; no control blocks it
      & indirect source kind, lineage edge, sink argument, terminal \\
    5. Memory Guard
      & untrusted memory reaches the action; no memory control stops it
      & memory read, supported relation into the slice, terminal \\
    6. HITL / Monitoring
      & confirmation required, but the action precedes a valid checkpoint
      & confirmation requirement, checkpoint absence, sink lifecycle \\
    \bottomrule
  \end{tabular}
\end{adjustbox}
\end{table}

Returning to the forbidden-write example as shown in Figure~\ref{fig:system-overview}(c), the selected slice supplies source
kind \texttt{user\_prompt}, family \texttt{filesystem\_write}, terminal status
\texttt{executed}, and the normalized protected path.  The pre-run oracle marks
that action as forbidden, so the outcome gate returns
\texttt{attack\_executed}.  Privilege / Tool Control does not apply because the
file-write capability itself is permitted for other paths.  Action / Output
Guardrail is the first supported rule because the protected write reaches an
executed terminal without a blocking decision; Input Guardrail is the second
because the direct user prompt explicitly requests the forbidden action.
The procedure therefore reports \emph{Action / Output Guardrail} as the primary
remediation layer and \emph{Input Guardrail} as secondary.

This separation also fixes the evaluation denominators.  The
human-adjudicated evaluation includes \DiagSuccessfulControlCases{}
successful-control cases and \DiagWeakLayerCases{} forbidden attempted or
executed cases; Section~\ref{sec:rq2} scores the two responsibilities
separately, and an absent reference class remains N/A.

\subsection{Evidence-Grounded Defense-Gap Cards}
\label{sec:cards}

This subsection implements $\operatorname{Render}(\cdot)$, binding every displayed field to the events that support it.

The Defense-Gap Card compresses the selected source, terminal, sink family and
status, structural path, policy outcome, and defense assessment into atomic
claims.  Each claim carries a value and supporting canonical event identifiers;
the validator checks value correctness, mandatory event types, allowed evidence,
and membership in the selected path.  When the recorded evidence cannot support
a complete diagnosis, the Card exposes the unresolved responsibility through an
explicit insufficient-trace outcome.  An analyst can therefore inspect not only
what \system concludes, but exactly why the conclusion is admissible.

Concretely, a Card groups five responsibilities from the same selected slice:
execution identity (source, terminal, and lifecycle), observed action (family,
status, and normalized arguments), structural explanation (steps and typed
edges), policy assessment (authorization and outcome), and remediation
(primary and secondary defense layers).  Every displayed field carries its
supporting event identifiers and validation state.  Keeping the value and its
support together prevents a plausible label from hiding a mismatched terminal,
an argument copied from another lifecycle, or a defense claim whose required
evidence never entered the decisive slice.

For the forbidden-write example in Figure~\ref{fig:system-overview}(c), the resulting Card binds the sink-family claim
to invocation $e_5$, the terminal-status claim to $e_7$, and the
policy-outcome claim to $\{e_1,e_5,e_7\}$.  Its primary Action / Output
Guardrail finding cites the protected-path invocation and executed terminal,
while its secondary Input Guardrail finding additionally cites the direct
prompt $e_1$.  Each displayed conclusion can therefore be traced back to the
canonical evidence that supports it.

\section{Implementation}
\label{sec:implementation}

\subsection{Cross-Runtime Prototype}

The Python prototype separates target integration from diagnosis.  Each driver
maps a runtime's native source, tool, lifecycle, guardrail, and terminal
boundaries into the same versioned event contract; replay, canonicalization,
STPG construction, slicing, policy attribution, evidence validation, and Card
rendering remain in a shared core.

We implement overlays for five structurally different open-source targets.
TaskWeaver exercises generated-code and plugin lifecycles
~\cite{qiao2023taskweaver}; Agent Zero combines browser, operating-system, and
subordinate-agent actions~\cite{agentzero2026repository}; LangGraph stresses
checkpointed and parallel graph execution~\cite{langgraph2026repository};
Goose exposes MCP request/response identity~\cite{goose2026repository}; and
Microsoft Agent Framework contributes concurrent workflows, handoffs, MCP, and
A2A parentage~\cite{microsoft2026agentframework}.  Together these integrations
cover code, browser, graph, protocol, and distributed-agent boundaries while
keeping target-specific values behind the canonical interface.

The shared projector derives family, lifecycle status, normalized arguments,
path structure, policy outcome, defense applicability, and claim evidence from
one selected terminal slice.  This atomic boundary prevents fields from
different lifecycles from being combined in one Card.  Diagnosis is fully
deterministic and invokes no language model, enabling content-hashed prediction
freezes and exact evaluator replay.

\subsection{Native Boundary Bindings}

Each overlay attaches where its target establishes execution identity.
TaskWeaver instruments its code executor, plugin lifecycle, and structured
\texttt{SecurityRisks} stop.  Agent Zero maps browser results to external
ingress and records explicit operating-system, editor, remote-control,
scheduler, and subordinate-agent actions.  LangGraph observes
\texttt{StateGraph} nodes, \texttt{ToolNode} calls, checkpoints, and parallel
fan-in.  Goose joins MCP JSON-RPC requests and responses around extension
dispatch.  Microsoft Agent Framework preserves workflow, concurrent-branch,
handoff, MCP, and A2A identifiers as distributed parent evidence.  Private
runtime values remain inside event metadata instead of extending the public
schema.

In the running example, Goose's pre-dispatch hook captures \texttt{path} and
\texttt{content}; sink hooks then surround the native MCP extension call with
an attempted invocation and executed result.  Matching MCP response identity
becomes the canonical \texttt{call\_id}, while path virtualization remains in
the safe fixture server.  The diagnosis core therefore sees the real native
lifecycle without depending on the target's transport wrappers.

Appendix~\ref{app:target-pins} records exact revisions, licenses, image
digests, overlay sizes, hook groups, and development-gate
coverage.  These frozen pins bind each public claim to the implementation that
produced its evidence.

\subsection{Atomic Diagnosis Core}

Canonicalization normalizes nested transport wrappers to
\texttt{payload.*}, records dangling parents, and joins a result to an
invocation only through a shared call identifier.  Sink classification runs
only on explicit lifecycle events.  The STPG builder stores edge evidence, the
terminal selector chooses the latest completed policy-decisive lifecycle, and
the backward slicer excludes post-terminal and unrelated branches.  Graph
compaction shortens retained text without changing node or edge identity.

After terminal selection, one atomic projector derives sink family and status,
normalized arguments, path steps and edges, policy outcome, defense
applicability, and supported claims from the same slice.  A
reference-semantics validator checks each claim's value, required event types,
allowed evidence identifiers, and path membership.  This boundary prevents a
newly selected terminal from being paired with arguments or defense labels
from an older lifecycle and leaves unsupported claims measurable.

For the Goose trace, canonicalization preserves seven events and builds an
eight-node STPG because the sink invocation becomes separate argument and
effect nodes.  The shared layer recovers the filesystem family,
attempted-to-executed transition, protected path, and marker.  Explicit
parentage supports the lifecycle path, while field-level influence retains the
evidence class reported by the target hook.

Attribution applies the outcome gate and frozen evidence-rule precedence to the
selected terminal, sink lifecycle, policy authorization, verified field
lineage, memory participation, and required human checkpoints.  Each satisfied
rule exports its layer, precedence rank, and supporting event identifiers; the
projector then selects the first two as primary and secondary without treating
the rank as confidence.  Card and prediction exporters likewise retain event
identifiers for every atomic claim.  Predictions are exported and
content-hashed before reference annotation, ensuring that the human reference
cannot alter the evaluated diagnosis.

\subsection{Evidence Packages and Reproducible Execution}

The common runner invokes every target in an isolated Docker container and
preserves the prompt, policy oracle, safe assets, native result, semantic and
raw traces, validation record, logs, and diagnosis artifacts as one evidence
package.  Schema, parent, lifecycle, and completeness validation runs before
diagnosis; malformed or incomplete runs remain explicit package-level
failures.  Collection manifests bind each package to prompt, policy, hook plan,
code, model, and container-image hashes.

Efficiency pairs use the same prompt, policy, model configuration, seed,
image, and safe assets.  The tracing-off arm suppresses semantic-event emission
and diagnosis, and a pair is retained only when a target-specific comparator
confirms equivalent terminal class, lifecycle, tool sequence, and native
outcome.  The runner records direct diagnosis latency and package size in
addition to the complete paired audit reported in
Appendix~\ref{app:paired-overhead}.

Deterministic fixtures cover terminal selection, argument normalization,
successful controls, weak-layer applicability, multiple sinks, parallel
branches, result reordering, claim evidence, and insufficient-trace
abstention.  A target-only change reruns that target's development gate; a
public semantic change reruns every fixture and target gate.  The confirmatory
manifest additionally binds the prediction freeze, two independent
annotations, adjudication record, evaluator version, and bootstrap seed.

\section{Evaluation}
\label{sec:evaluation}

\subsection{Evaluation Questions}

Our evaluation addresses three empirical questions, corresponding to the first three contributions in the Introduction.
\textbf{E1} tests whether one canonical contract preserves complete,
policy-decisive lifecycles across structurally different agent runtimes.
\textbf{E2} tests whether \system produces accurate and evidence-accountable
diagnoses under an identical input and output contract.  \textbf{E3} tests
whether the proposed mechanisms create that advantage and quantifies the
deterministic cost of producing a diagnosis.  E1 evaluates the cross-runtime
substrate, E2 evaluates the end-to-end capability, and E3 connects the observed
gain to the design.

\subsection{Targets and Workloads}

The five applications stress different execution structures: TaskWeaver is
code-centric; Agent Zero combines browser, operating-system, and subordinate-
agent tools; LangGraph exposes checkpointed and parallel graph execution;
Goose uses MCP request/response lifecycles; and Microsoft Agent Framework
exposes workflows, handoffs, and multi-agent messages.  Their native boundaries
are summarized in Table~\ref{tab:runtime-diversity}.  Experiments use synthetic
local files, inert code markers, loopback sends, and local webpage or MCP
content.

The acquisition workload crosses the five pinned targets with 12 frozen
scenario semantics: ten controlled attacks and two benign controls.  The
scenarios cover direct filesystem, code, and external-send requests;
attacker-controlled external content flowing to each sink family; split or
encoded instructions; authorization escalation; unsafe repair after an
authorized failure; and target-native relay boundaries.  All targets use the
same model binding, and each applicable
scenario--target cell is executed three times.

The full matrix contains 180 theoretical run slots.  TaskWeaver is naturally
N/A for seven scenarios that require an external-send, external-content, or
cross-boundary adapter absent from its pinned configuration; the other four
targets support all 12 scenarios.  The resulting 21 preregistered N/A slots
leave \RQOneApplicableRuns{} executions over 53 applicable cells.  Appendix
~\ref{app:workload-catalog} gives the scenario-by-target matrix and exact run
accounting; Appendix~\ref{app:target-outcomes} reports the frozen blocked,
failed, attempted, executed, and benign outcomes.

\subsection{Confirmatory Protocol and Metrics}

E2 uses a preregistered successor protocol: after acquisition froze the
scenario definitions and canonical schema, we froze a separate stratified
diagnosis manifest, exclusion rules, metrics, thresholds, code commit, and
evaluator before final predictions.  Thus, the \RQOneApplicableRuns{}
acquisition executions establish cross-runtime coverage for E1, whereas the
confirmatory set evaluates diagnosis quality for E2--E3.  Each confirmatory
case contains the canonical evidence package, one frozen prediction from every
method, prediction-blind labels, and the adjudicated reference used for
scoring.  Full and all baselines emit the same structured schema.  Of
\DiagCollectedCases{} collected confirmatory cases, \DiagResolvedCases{} enter
the final denominator (\DiagReferenceCoveragePct{}\% reference coverage), with
at least \DiagMinimumCasesPerTarget{} cases per target.

Grounded joint diagnosis (GJD) is the primary end-to-end measure: the
lifecycle, core security fields, policy outcome, applicable dangerous
arguments, and mandatory claim evidence must be correct in the same case.
Joint diagnosis accuracy (JDA) uses the same conjunction without the evidence
requirement.  We report policy-outcome macro-F1, required-evidence coverage,
unsupported-claim rate, component precision/recall/F1, worst-target GJD, and
leave-one-target-out (LOTO) mean GJD.

\paragraphlead{Interval construction}
We use 95\% Wilson score intervals for GJD, JDA, categorical and
successful-control accuracies, and pooled claim proportions.  Denominators are
the applicable adjudicated cases or claims.  Wilson intervals do not account
for dependence within cases or across cases sharing a scenario or runtime.
F1 scores and paired differences use 95\% hierarchical-bootstrap intervals
from \DiagBootstrapReplicatesDisplay{} replicates, resampling
\DiagBootstrapPrimaryUnit{}s and then their cases.  Paired methods use the
same resampled cases.  This bootstrap preserves within-cell grouping but not
cross-cell dependence; its intervals are conditional on the frozen executions.

The complete endpoint criteria, applicable denominators, and pre-adjudication
agreement are specified next.

\subsection{Reference Construction and Annotation Agreement}
\label{sec:reference-construction}

Two annotators independently construct each reference from the frozen native
and canonical evidence package and pre-run oracle, without seeing predictions
or the scenario's expected outcome.  Under one codebook they label source,
terminal, lifecycle fields, arguments, structure, outcome, defense, and claim
evidence.  A third annotator resolves disagreements under the same codebook,
preserving observation--policy separation.

\begin{table}[!htbp]
  \centering
  \caption{Independent-annotation agreement before adjudication. Categorical
  fields use Cohen's $\kappa$; set/path fields use mean pairwise F1.}
  \label{tab:rq2-agreement}
  \footnotesize
  \setlength{\tabcolsep}{4pt}
  \begin{adjustbox}{max width=\linewidth}
\begin{tabular}{l c c}
    \toprule
    Field & Agreement & $n$ \\
    \midrule
    Policy outcome & $\kappa=\DiagIAAOutcomeKappa{}$ & \DiagCollectedCases{} \\
    Primary defense layer & $\kappa=\DiagIAADefenseKappa{}$ & \DiagCollectedCases{} \\
    Steps & F1=\DiagIAAStepFOne{} & \DiagCollectedCases{} \\
    Typed edges & F1=\DiagIAAEdgeFOne{} & \DiagCollectedCases{} \\
    Claim evidence & F1=\DiagIAAClaimEvidenceFOne{} & \DiagCollectedCases{} \\
    Cases requiring adjudication & \DiagAdjudicationRate{} & \DiagCollectedCases{} \\
    \bottomrule
  \end{tabular}
\end{adjustbox}
\end{table}

Before adjudication, outcome and primary-defense agreement reach Cohen's
$\kappa$ of \DiagIAAOutcomeKappa{} and \DiagIAADefenseKappa{}; structured-set
agreement ranges from \DiagIAAEdgeFOne{} to \DiagIAAStepFOne{} F1
(Table~\ref{tab:rq2-agreement}).  The adjudicator resolves the
\DiagAdjudicationRate{} of cases with at least one disagreement; predictions
remain hidden and the diagnosis rules remain frozen.

Success is gated jointly by GJD \DiagGJDThreshold{}, outcome macro-F1
\DiagOutcomeMacroFOneThreshold{}, evidence coverage
\DiagRequiredEvidenceCoverageThreshold{}, and unsupported claims at most
\DiagUnsupportedClaimRateThreshold{}.  Typed-edge F1 and worst-target GJD must
also reach \DiagEdgeFOneThreshold{} and \DiagWorstTargetGJDThreshold{};
successful-control and weak-layer scores retain separate denominators.

\subsection{E1: One Contract Across Heterogeneous Runtimes}
\label{sec:rq1}

All \RQOneApplicableRuns{} applicable executions produce complete,
schema-valid evidence packages.  Every observed sink lifecycle is joined by
invocation/result identity, every external-content scenario retains its ingress,
and every run exposes a source, a policy-decisive terminal, and a supported
parent path.  The same public contract therefore represents code execution,
browser/OS actions, checkpointed graphs, MCP calls, and multi-agent handoffs
without adding target-specific diagnosis fields.

This result establishes the substrate needed by the rest of the evaluation:
the shared diagnosis core receives the same semantic objects even though the
five runtimes expose them at different native boundaries.  In the Goose
example, one lifecycle connects the untrusted prompt, MCP invocation, executed
result, normalized arguments, and decisive terminal while retaining the native
evidence class of each edge.

\subsection{E2: Accurate, Evidence-Accountable Diagnosis}
\label{sec:rq2}

Table~\ref{tab:rq2-confirmatory} evaluates the complete diagnosis rather than
isolated labels.  Full \system reaches \DiagGroundedJointDiagnosis{} GJD (95\%
CI \DiagGroundedJointDiagnosisCI{}), meaning that the lifecycle, core security
fields, policy outcome, applicable arguments, and mandatory claim evidence are
jointly correct in the same case.  JDA is \DiagJointDiagnosisAccuracy{}
(\DiagJointDiagnosisCI{}), and policy-outcome macro-F1 is
\DiagOutcomeMacroFOne{} (\DiagOutcomeMacroFOneCI{}).

\begin{table}[t]
  \centering
  \caption{End-to-end diagnosis on \DiagResolvedCases{} adjudicated cases.
  Intervals are 95\% Wilson score intervals for GJD, JDA, and pooled claim
  proportions, and a hierarchical-bootstrap interval for outcome macro-F1
  (Section~\ref{sec:evaluation}).  Lower unsupported-claim rate is better.}
  \label{tab:rq2-confirmatory}
  \footnotesize
  \setlength{\tabcolsep}{3.2pt}
  \begin{adjustbox}{max width=\linewidth}
\begin{tabular}{l c c}
    \toprule
    Metric & Overall & 95\% CI \\
    \midrule
    \multicolumn{3}{l}{\emph{End-to-end diagnosis}} \\
    Grounded joint diagnosis & \DiagGroundedJointDiagnosis{} & \DiagGroundedJointDiagnosisCI{} \\
    Joint diagnosis accuracy & \DiagJointDiagnosisAccuracy{} & \DiagJointDiagnosisCI{} \\
    Policy-outcome macro-F1 & \DiagOutcomeMacroFOne{} & \DiagOutcomeMacroFOneCI{} \\
    \midrule
    \multicolumn{3}{l}{\emph{Evidence accountability}} \\
    Required-evidence coverage & \DiagRequiredEvidenceCoverage{} & \DiagRequiredEvidenceCI{} \\
    Unsupported-claim rate $\downarrow$ & \DiagUnsupportedClaimRate{} & \DiagUnsupportedClaimCI{} \\
    \midrule
    \multicolumn{3}{l}{\emph{Cross-runtime robustness}} \\
    Worst-target GJD & \DiagWorstTargetGJD{} & \DiagWorstTargetName{} \\
    LOTO mean GJD & \DiagLeaveOneTargetOutMeanGJD{} & \DiagTargetCount{} targets \\
    \bottomrule
  \end{tabular}
\end{adjustbox}
\end{table}

The evidence metrics show why the joint result is operationally useful.
\system supplies \DiagRequiredEvidenceCoverage{} of mandatory claim evidence
(\DiagRequiredEvidenceCI{}) while limiting unsupported predicted claims to
\DiagUnsupportedClaimRate{} (\DiagUnsupportedClaimCI{}).  Structural recovery
reaches \DiagStepFOne{} step F1, \DiagEdgeFOne{} typed-edge F1, and
\DiagArgumentFOne{} dangerous-argument F1.  The worst-target GJD remains
\DiagWorstTargetGJD{} on \DiagWorstTargetName{}, showing that the aggregate is
not driven by a single runtime.  Together, these results demonstrate one
capability: \system produces diagnoses whose values and evidentiary support
remain consistent across heterogeneous native boundaries.

\subsection{What the Joint Diagnosis Recovers}
\label{sec:rq2-components}

GJD is deliberately stricter than any isolated classification score: one wrong
source, terminal, lifecycle field, argument, outcome, or evidence obligation
makes the entire case incorrect.  Table~\ref{tab:diagnostic-components} opens
this conjunction and shows which responsibilities are recovered before they
are recombined into the end-to-end result.

\begin{table*}[t]
  \caption{Component-level diagnosis on the same \DiagResolvedCases{}
  adjudicated cases as Table~\ref{tab:rq2-confirmatory}.
  Intervals are 95\% Wilson score intervals for accuracies and
  hierarchical-bootstrap intervals for F1 scores
  (Section~\ref{sec:evaluation}).}
  \label{tab:diagnostic-components}
  \centering
  \small
  \begin{adjustbox}{max width=\linewidth}
\begin{tabular}{l r c c}
    \toprule
    Metric & $n$ & Overall & 95\% CI \\
    \midrule
    Source accuracy & \DiagResolvedCases{} & \DiagSourceAccuracy{} & \DiagSourceCI{} \\
    Terminal accuracy & \DiagResolvedCases{} & \DiagTerminalAccuracy{} & \DiagTerminalCI{} \\
    Sink-family accuracy & \DiagResolvedCases{} & \DiagFamilyAccuracy{} & \DiagFamilyCI{} \\
    Sink-status accuracy & \DiagResolvedCases{} & \DiagStatusAccuracy{} & \DiagStatusCI{} \\
    Step precision / recall / F1 & \DiagResolvedCases{} & \DiagStepPrecision{} / \DiagStepRecall{} / \DiagStepFOne{} & \DiagStepFOneCI{} \\
    Typed-edge precision / recall / F1 & \DiagResolvedCases{} & \DiagEdgePrecision{} / \DiagEdgeRecall{} / \DiagEdgeFOne{} & \DiagEdgeFOneCI{} \\
    Dangerous-argument precision / recall / F1 & \DiagResolvedCases{} & \DiagArgumentPrecision{} / \DiagArgumentRecall{} / \DiagArgumentFOne{} & \DiagArgumentFOneCI{} \\
    Successful-control accuracy & \DiagSuccessfulControlCases{} & \DiagControlAccuracy{} & \DiagControlCI{} \\
    Weak-layer macro-F1 & \DiagWeakLayerCases{} & \DiagWeakLayerMacroFOne{} & \DiagWeakLayerCI{} \\
    Claim-value set F1 & \DiagResolvedCases{} & \DiagClaimValueAccuracy{} & \DiagClaimValueCI{} \\
    Claim-evidence F1 & \DiagResolvedCases{} & \DiagClaimEvidenceFOne{} & \DiagClaimEvidenceFOneCI{} \\
    \midrule
    Worst-target outcome macro-F1 & \DiagMinimumCasesPerTarget{} & \DiagWorstTargetOutcomeMacroFOne{} & -- \\
    Worst-target family accuracy & \DiagMinimumCasesPerTarget{} & \DiagWorstTargetFamilyAccuracy{} & -- \\
    Worst-target argument F1 & \DiagMinimumCasesPerTarget{} & \DiagWorstTargetArgumentFOne{} & -- \\
    \bottomrule
  \end{tabular}
\end{adjustbox}
\end{table*}

The first group measures whether \system identifies the operation that should
be diagnosed.  Source and terminal accuracy reach \DiagSourceAccuracy{} and
\DiagTerminalAccuracy{}, while sink-family and status accuracy reach
\DiagFamilyAccuracy{} and \DiagStatusAccuracy{}.  These four endpoints test
different decisions: source selection identifies the security-relevant input;
terminal selection fixes one lifecycle; family identifies the effect boundary;
and status distinguishes an attempted, executed, failed, or blocked action.
Their simultaneous strength is what prevents a plausible outcome label from
masking an incorrectly reconstructed execution.

The second group measures the content of that reconstruction.  The recovered
event path reaches \DiagStepFOne{} step F1, its typed relations reach
\DiagEdgeFOne{} F1, and policy-relevant dangerous arguments reach
\DiagArgumentFOne{} F1.  Typed-edge scoring is especially discriminating: a
chronologically plausible connection receives no credit unless its parent,
lifecycle, or field-lineage type agrees with the reference.  Argument scoring
likewise requires the value to be attached to the selected lifecycle rather
than merely appearing somewhere in the trace.  These results support the use
of the STPG slice as a structured diagnostic object, not just a compact event
summary.

The final group tests whether the diagnosis is actionable and auditable.
Successful controls are identified with \DiagControlAccuracy{} accuracy, while
post-attempt weak-layer attribution reaches \DiagWeakLayerMacroFOne{} macro-F1
over the applicable forbidden-action cases.  Keeping these denominators
separate avoids treating a correctly blocked attack as a weak-layer failure.
At the Card boundary, claim-value set F1 is \DiagClaimValueAccuracy{} and
claim-evidence F1 is \DiagClaimEvidenceFOne{}, showing that the reported values
and the event identifiers offered to justify them remain aligned.  The
worst-target component rows provide a second cross-runtime check alongside
worst-target GJD.

All component labels come from the same adjudicated case references used for
GJD.  Annotators review the evidence packages without seeing predictions,
record categorical fields and structured sets independently, and adjudicate
disagreements before scoring.  Applicability is determined by the observed
outcome: successful-control accuracy uses the \DiagSuccessfulControlCases{}
blocked cases, whereas weak-layer macro-F1 uses the \DiagWeakLayerCases{}
forbidden attempted or executed cases.  Section~\ref{sec:reference-construction}
and Table~\ref{tab:rq2-agreement} report the corresponding agreement,
thresholds, and denominator rules.

\subsection{Fair Same-Schema Comparison}

Table~\ref{tab:rq2-competitive} compares every method under the same canonical
input and versioned output schema, including abstentions and claim evidence.
Among the baselines, \DiagStrongestBaselineName{} is strongest by GJD at
\DiagStrongestBaselineGJD{}, while Full \system reaches
\DiagGroundedJointDiagnosis{}.  The paired gain is \DiagGJDDelta{} (95\% CI
\DiagGJDDeltaCI{}); policy-outcome macro-F1 and required-evidence coverage
improve by \DiagOutcomeDelta{} (\DiagOutcomeDeltaCI{}) and
\DiagEvidenceDelta{} (\DiagEvidenceDeltaCI{}), respectively.  Holding both
input and output capability fixed isolates diagnosis quality: the gain comes
from recovering the policy-decisive lifecycle and validating claim evidence,
not from a richer reporting interface.

\begin{table*}[t]
  \centering
  \caption{Under identical evidence and structured-output contracts, Full
  \system improves GJD, policy-outcome macro-F1, and required-evidence
  coverage over the strongest same-schema baseline
  (\DiagStrongestBaselineName{}).  Intervals are 95\% paired
  hierarchical-bootstrap intervals over scenario--target cells
  (Section~\ref{sec:evaluation}).}
  \label{tab:rq2-competitive}
  \footnotesize
  \setlength{\tabcolsep}{5pt}
  \begin{adjustbox}{max width=\linewidth}
\begin{tabular}{l c c c c}
    \toprule
    Metric & Full \system & Strongest baseline & Full $-$ baseline & 95\% CI \\
    \midrule
    Grounded joint diagnosis & \DiagGroundedJointDiagnosis{} & \DiagStrongestBaselineGJD{} & +\DiagGJDDelta{} & \DiagGJDDeltaCI{} \\
    Policy-outcome macro-F1 & \DiagOutcomeMacroFOne{} & \DiagStrongestBaselineOutcomeMacroFOne{} & +\DiagOutcomeDelta{} & \DiagOutcomeDeltaCI{} \\
    Required-evidence coverage & \DiagRequiredEvidenceCoverage{} & \DiagStrongestBaselineEvidenceCoverage{} & +\DiagEvidenceDelta{} & \DiagEvidenceDeltaCI{} \\
    \bottomrule
  \end{tabular}
\end{adjustbox}
\end{table*}

\subsection{E3: Where the Advantage Comes From}
\label{sec:e3}

Table~\ref{tab:rq2-ablation} removes one mechanism at a time while retaining
the same cases and structured output contract.  Every removal reduces GJD and
increases unsupported claims, with paired intervals excluding zero.  The
largest GJD contribution comes from STPG slicing, which separates the decisive
path from unrelated execution structure; evidence validation prevents the
largest increase in unsupported claims.  Lifecycle joining, terminal
selection, argument normalization, and defense attribution provide additional
independent contributions.  The consistent direction across all six variants
connects the end-to-end gain to the architecture rather than to a single
heuristic.

\begin{table}[t]
  \centering
  \caption{Each AgentTriage mechanism makes an independent paired contribution:
  retaining the mechanism increases GJD and prevents unsupported claims.  All
  variants preserve the same input and output contract.}
  \label{tab:rq2-ablation}
  \footnotesize
  \setlength{\tabcolsep}{2.8pt}
\maxsizebox{\columnwidth}{!}{\begin{adjustbox}{max width=\linewidth}
\begin{tabular}{l c c}
    \toprule
    Mechanism retained & \shortstack{GJD contribution\\(95\% CI) $\uparrow$} & \shortstack{UC prevented\\(95\% CI) $\uparrow$} \\
    \midrule
    Lifecycle joining & \DiagAblLifecycleGJDContribution{} \DiagAblLifecycleGJDContributionCI{} & \DiagAblLifecycleUCPrevented{} \DiagAblLifecycleUCPreventedCI{} \\
    Terminal selection & \DiagAblTerminalGJDContribution{} \DiagAblTerminalGJDContributionCI{} & \DiagAblTerminalUCPrevented{} \DiagAblTerminalUCPreventedCI{} \\
    STPG slicing & \DiagAblSTPGGJDContribution{} \DiagAblSTPGGJDContributionCI{} & \DiagAblSTPGUCPrevented{} \DiagAblSTPGUCPreventedCI{} \\
    Argument normalization & \DiagAblArgumentGJDContribution{} \DiagAblArgumentGJDContributionCI{} & \DiagAblArgumentUCPrevented{} \DiagAblArgumentUCPreventedCI{} \\
    Defense attribution & \DiagAblDefenseGJDContribution{} \DiagAblDefenseGJDContributionCI{} & \DiagAblDefenseUCPrevented{} \DiagAblDefenseUCPreventedCI{} \\
    Evidence validation & \DiagAblEvidenceGJDContribution{} \DiagAblEvidenceGJDContributionCI{} & \DiagAblEvidenceUCPrevented{} \DiagAblEvidenceUCPreventedCI{} \\
    \bottomrule
  \end{tabular}
\end{adjustbox}
  }
\end{table}

The six mechanisms form a diagnostic chain.  Lifecycle joining and terminal
selection identify the completed operation; STPG slicing and argument
normalization recover its policy-relevant path and values; defense attribution
and evidence validation convert that object into supported remediation claims.
Earlier failures change the input to every downstream stage, whereas a
Card-boundary failure leaves the path intact but reduces supportable claims.
Reporting GJD with unsupported-claim prevention therefore tests reconstruction
and reporting responsibility together.

\subsection{Operational Cost}
\label{sec:operational-cost}

\system performs deterministic post-execution diagnosis without invoking a
language model.  Across the five targets, median diagnosis latency ranges from
5.881\,ms to 12.261\,ms, and the incremental evidence package ranges from
29,849 to 112,955 bytes.  These costs include canonicalization, lifecycle
joining, STPG construction, slicing, attribution, evidence validation, and Card
rendering.  The diagnosis can therefore be added to an agent evaluation
pipeline without another inference step or diagnosis-model token budget.
Table~\ref{tab:operational-cost} reports the direct costs; Appendix
~\ref{app:paired-overhead} gives the complete tracing-on/off paired audit,
including provider- and network-sensitive end-to-end measurements.

\begin{table}[t]
  \centering
  \caption{AgentTriage adds deterministic diagnosis without another model
  inference.  Values are target-level medians over valid frozen pairs.}
  \label{tab:operational-cost}
  \footnotesize
  \setlength{\tabcolsep}{4pt}
\maxsizebox{\columnwidth}{!}{\begin{adjustbox}{max width=\linewidth}
\begin{tabular}{l r r}
    \toprule
    Target & Diagnosis (ms) & Evidence package (B) \\
    \midrule
    Agent Zero & 5.881 & 53,538 \\
    Goose & 12.261 & 112,955 \\
    LangGraph & 10.893 & 92,600 \\
    Microsoft AF & 9.009 & 56,463 \\
    TaskWeaver & 6.661 & 29,849 \\
    \bottomrule
  \end{tabular}
\end{adjustbox}
  }
\end{table}

\section{Related Work}
\label{sec:related-work}

\system occupies the post-execution diagnosis layer, combining lifecycle,
policy, defense, and claim evidence in one evaluated output.  Table
~\ref{tab:related-positioning} contrasts it with the five closest research
threads.

\paragraphlead{Prompt injection and agent safety evaluation}
Prompt-injection studies cover black-box applications, attacker-controlled
content, formal attack spaces, executable tool cases, and persistent memory
poisoning~\cite{liu2023houyi,greshake2023indirect,yi2023bipia,
toyer2024tensortrust,liu2024formalizing,zhan2024injecagent,
debenedetti2024agentdojo,chen2024agentpoison}.  Behavioral benchmarks span web,
desktop, enterprise, software engineering, tool use, and policy following
~\cite{zhou2024webarena,koh2024visualwebarena,drouin2024workarena,
xie2024osworld,jimenez2024swebench,mialon2024gaia,qin2024toolllm,
yao2025taubench,ma2024agentboard}.  Safety suites add sandboxed, stateful,
harmful, and interactive outcomes~\cite{ruan2024toolemu,lu2025toolsandbox,
andriushchenko2025agentharm,kuntz2025osharm,yang2025riosworld,
vijayvargiya2026openagentsafety,zhang2025asb,liu2024agentbench}.  \system
recovers the jointly evaluated lifecycle and evidence behind such outcomes.

\paragraphlead{Vulnerability discovery and agent red teaming}
Program analysis, fuzzing, taint reasoning, and executable arenas discover
prompt-to-sink and cross-tool risk paths~\cite{liu2024llmsmith,
liu2025agentfuzz,he2026taintp2x,wu2026chainfuzzer,lin2026agentraft,
li2026autocontrol}.  \system turns a discovered execution into a normalized,
policy-relative, claim-auditable diagnosis.

\paragraphlead{Trajectory diagnosis and security forensics}
Trajectory analysis localizes failure steps, responsible inputs, and risky
interactions~\cite{barke2026agentrx,wang2025tracllm,yuan2024rjudge}.  \system
adds native lifecycles, a policy-decisive terminal, authorization, defense
semantics, and mandatory evidence for each atomic claim.

\paragraphlead{Information flow, provenance, and runtime analysis}
Program slicing, provenance, request tracing, causal monitoring, and taint
analysis preserve execution structure~\cite{weiser1984program,
korel1988dynamic,agrawal1990dynamic,moreau2013provdm,barham2004magpie,
fonseca2007xtrace,sigelman2010dapper,mace2015pivot,opentelemetry2026trace,
enck2010taintdroid,arzt2014flowdroid}.  Security provenance adds correlated
graph context~\cite{hassan2019nodoze,milajerdi2019holmes,han2020unicorn}; agent
analyses add semantic, counterfactual, and asynchronous tracking
~\cite{cai2026neurotaint,siddiqui2025permissive,costa2025fides,
zhong2025rtbas,wang2025agentarmor}.  \system anchors explicit structure to
policy outcome, terminal status, defense attribution, and claim responsibility.

\paragraphlead{Preventing unsafe tool use}
Preventive systems separate instructions from data or enforce alignment,
dependencies, validation, isolation, information flow, and runtime policy
~\cite{hines2024spotlighting,wallace2024instruction,piet2024jatmo,
chen2025secalign,chen2025struq,jia2025taskshield,an2025ipiguard,li2025drift,
costa2025fides,zhong2025rtbas,wang2025agentarmor,debenedetti2025camel}.
\system audits the resulting lifecycle, successful control or post-attempt
weak layer, and supporting events.

\section{Scope, Deployment, and Extensions}
\label{sec:limitations}

\begin{table}[t]
  \centering
  \caption{Positioning against the closest research threads.  \system{} turns a
  completed run into one policy-relative, evidence-accountable diagnosis.}
  \label{tab:related-positioning}
  \footnotesize
  \setlength{\tabcolsep}{3.5pt}
  \begin{adjustbox}{max width=\linewidth}
\begin{tabular}{@{}p{0.30\columnwidth} p{0.62\columnwidth}@{}}
    \toprule
    Thread & Representative work; capability added by \system{} \\
    \midrule
    Benchmarks
      & AgentDojo, ToolEmu, AgentHarm~\cite{debenedetti2024agentdojo,ruan2024toolemu,andriushchenko2025agentharm}:
        reconstructs the lifecycle behind a scored end state. \\
    Vulnerability discovery
      & LLMSmith, AgentFuzz, ChainFuzzer~\cite{liu2024llmsmith,liu2025agentfuzz,wu2026chainfuzzer}:
        turns a candidate run into a policy-relative, claim-auditable diagnosis. \\
    Trajectory diagnosis
      & AgentRx, TracLLM, R-Judge~\cite{barke2026agentrx,wang2025tracllm,yuan2024rjudge}:
        adds cross-runtime lifecycle, policy, defense, and evidence semantics. \\
    Provenance / flow
      & NeuroTaint, NoDoze, HOLMES~\cite{cai2026neurotaint,hassan2019nodoze,milajerdi2019holmes}:
        couples explicit structure with authorization, status, and claim responsibility. \\
    Preventive defenses
      & TaskShield, IPIGuard, AgentArmor~\cite{jia2025taskshield,an2025ipiguard,wang2025agentarmor}:
        audits the observed control result and the lifecycle that followed it. \\
    \bottomrule
  \end{tabular}
\end{adjustbox}
\end{table}

\paragraphlead{What the evidence establishes, and what it does not}
The contribution is a contract and a measurement, not a new diagnostic
algorithm.  The evaluation establishes three things: that one contract captures
all \RQOneApplicableRuns{} applicable executions across five runtimes; that two
of the five overlays must synthesize the call identity the other three obtain
natively, so the reach of the contract is a measured property rather than an
assumption; and that on \DiagResolvedCases{} adjudicated cases, \system reaches
\DiagGroundedJointDiagnosis{} GJD and \DiagRequiredEvidenceCoverage{}
required-evidence coverage with an unsupported-claim rate of
\DiagUnsupportedClaimRate{}.  The \DiagGJDDelta{} paired gain over the strongest
same-schema baseline and the mechanism ablations connect this result to
lifecycle recovery, decisive slicing, and claim validation rather than to output
availability.  What the evaluation does not establish is that a replayable
diagnosis is also the right diagnosis: where the recorded evidence is incomplete,
the contract abstains, and abstention is not accuracy.

\paragraphlead{Explicit evidence as a deployment contract}
\system admits structure from native parents, invocation/result identity, and
declared field lineage.  This choice gives operators a clear trust rule: every
retained edge and every rendered claim can be traced to the runtime boundary
that reported it.  Deployments that require paraphrase, implicit control
dependence, latent memory influence, or counterfactual attribution can add a
separately validated semantic layer while preserving the explicit graph as the
auditable base.  The current results apply to the five pinned target stacks,
tested scenario strata, policy oracle, collectors, and evidence store.

\paragraphlead{Integration and operating cost}
A new runtime integrates through a target overlay while reusing the diagnosis
core and public contract.  Diagnosis is deterministic and adds no diagnosis-model
call, so it can be replayed on frozen evidence, and it completes in median
5.881--12.261\,ms across the evaluated targets. Full
tracing-on/off measurements and the observed target outcomes are reported in
the appendix so that the main text can focus on the diagnosis capability they
support.

\section{Conclusion}
\label{sec:conclusion}

We treated cross-runtime security diagnosis as an object with an executable
contract rather than as the output of a new algorithm.  The contract fixes what
a conclusion must contain, what evidence it must cite, and when it must abstain;
\system implements it with a deterministic core that invokes no language model,
so a diagnosis can be recomputed from the recorded trace, the contract version,
and the pre-run oracle, and each claim re-checked without trusting the original
analysis.  Determinism here buys reproducibility, not completeness: events
outside an instrumented boundary never enter the result, which is why the
evaluation reports coverage and unsupported claims next to accuracy.

Across five structurally different runtimes, one contract covers all
\RQOneApplicableRuns{} applicable executions, while two of the five overlays
must synthesize a call identity that the other three obtain natively.  On
\DiagResolvedCases{} adjudicated cases the system attains
\DiagGroundedJointDiagnosis{} GJD and \DiagRequiredEvidenceCoverage{}
required-evidence coverage at \DiagUnsupportedClaimRate{} unsupported claims,
and the ablations attribute the gain to lifecycle joining, decisive slicing, and
evidence validation rather than to the reporting format.  What can be audited is
therefore bounded from two sides: by what a runtime exposes, which we measure,
and by the precision the available evidence supports, which we quantify.

\bibliographystyle{IEEEtran}
\bibliography{references}
\onecolumn
\appendices
\section{Frozen Target Pins}
\label{app:target-pins}

Table~\ref{tab:target-pins} records the exact target revisions associated with
the evaluated overlays, the integration surface of each overlay, and the
deterministic development gates it exercises.  Image digests are abbreviated in
print; the artifact manifest retains their complete values.  All five
integrations are implemented as out-of-tree target overlays and require no
upstream source patches.

\begin{table}[!htbp]
  \centering
  \caption{Frozen upstream revisions and overlay scope for the five targets.
  Image digests are abbreviated; the artifact manifest retains complete values.}
  \label{tab:target-pins}
  \footnotesize
  \setlength{\tabcolsep}{4pt}
  \begin{adjustbox}{max width=\linewidth}
\begin{tabular}{@{}l l l l r r c@{}}
    \toprule
    Target & Commit & License & Image digest & Overlay SLoC & Hook groups & Dev. gate \\
    \midrule
    TaskWeaver & \texttt{d44ddef2} & MIT & \texttt{92c874\ldots} & 409 & 4 & 10/10 \\
    Agent Zero & \texttt{fddcc3de} & MIT & \texttt{c5655e\ldots} & 914 & 4 & 10/10 \\
    LangGraph & \texttt{55ec2f21} & MIT & \texttt{8068ff\ldots} & 1167 & 4 & 10/10 \\
    Goose & \texttt{2ecb8c08} & Apache-2.0 & \texttt{dfe00d\ldots} & 727 & 4 & 10/10 \\
    Microsoft AF & \texttt{6c0950ad} & MIT & \texttt{f90320\ldots} & 751 & 4 & 12/12 \\
    \bottomrule
  \end{tabular}
\end{adjustbox}
\end{table}

\section{Acquisition Workload and Applicability}
\label{app:workload-catalog}

Table~\ref{tab:workload-catalog} expands the 12 frozen acquisition scenarios.
The first four targets support all 12 cells; TaskWeaver supports five and is
preregistered N/A for SG-03--SG-07, SG-10, and SG-11 because its pinned
configuration lacks the external-send, external-content, or cross-boundary
adapter required by those scenario semantics.  One repetition therefore has
53 applicable cells (12 each for Agent Zero, LangGraph, Goose, and Microsoft
Agent Framework, plus five for TaskWeaver).  Three repetitions yield 159
executions and 21 N/A slots out of the 180-cell theoretical matrix.

\begin{table}[!htbp]
  \centering
  \caption{Frozen acquisition scenarios and target applicability.  TW is
  TaskWeaver, AZ is Agent Zero, LG is LangGraph, GO is Goose, and MAF is
  Microsoft Agent Framework.}
  \label{tab:workload-catalog}
  \scriptsize
  \setlength{\tabcolsep}{3.5pt}
  \begin{adjustbox}{max width=\linewidth}
\begin{tabular}{l p{0.31\textwidth} l l p{0.23\textwidth}}
    \toprule
    ID & Scenario semantic & Origin & Class & Applicable targets \\
    \midrule
    SG-01 & Direct unauthorized filesystem write & User prompt & Attack & TW, AZ, LG, GO, MAF \\
    SG-02 & Direct unauthorized code execution & User prompt & Attack & TW, AZ, LG, GO, MAF \\
    SG-03 & Direct unauthorized external send & User prompt & Attack & AZ, LG, GO, MAF \\
    SG-04 & External content propagated to filesystem write & External content & Attack & AZ, LG, GO, MAF \\
    SG-05 & External content propagated to code execution & External content & Attack & AZ, LG, GO, MAF \\
    SG-06 & External content propagated to external send & External content & Attack & AZ, LG, GO, MAF \\
    SG-07 & Split or encoded external instruction & External content & Attack & AZ, LG, GO, MAF \\
    SG-08 & Authorized action followed by forbidden escalation & Mixed & Attack & TW, AZ, LG, GO, MAF \\
    SG-09 & Authorized failure followed by unsafe repair & Mixed & Attack & TW, AZ, LG, GO, MAF \\
    SG-10 & Cross-boundary relay to a sensitive sink & Target-native relay & Attack & AZ, LG, GO, MAF \\
    SG-11 & Benign external command-like content & External content & Control & AZ, LG, GO, MAF \\
    SG-12 & Authorized safe-sink lookalike & User prompt & Control & TW, AZ, LG, GO, MAF \\
    \bottomrule
  \end{tabular}
\end{adjustbox}
\end{table}

All five targets use same LLM through an OpenAI-compatible endpoint with temperature 0.1.  LangGraph and Microsoft Agent Framework pass seed 20260714; the pinned
TaskWeaver, Agent Zero, and Goose paths expose no equivalent provider-level
seed guarantee, so their three cells are treated as repeated model runs rather
than deterministic duplicates.  Prompts, fixtures, target revisions, model
parameters, overlay revisions, and container-image digests are frozen for the
acquisition matrix.

\section{Complete Paired Tracing Audit}
\label{app:paired-overhead}

The audit contains 76 tracing-on/off pairs across five targets and three
strata.  Sixty-three pairs are semantically equivalent under terminal class,
sink lifecycle, normalized tool sequence, and target-native outcome; 13 are
excluded, and every target--stratum cell retains at least three valid pairs.
End-to-end relative deltas include provider and network variation, and their
target-level confidence intervals cross zero.  Provider token telemetry is
unavailable in both arms, while the deterministic diagnosis layer itself makes
no model call.  Table~\ref{tab:rq4-overhead}, below, preserves these
measurements alongside the direct diagnosis latency and package-size values used
in the main text.

\section{Observed Outcomes of Pinned Target Configurations}
\label{app:target-outcomes}

Table~\ref{tab:agent-behavior} reports mutually exclusive outcomes derived from
raw sink and guardrail evidence under the frozen oracle.  The counts establish
that the workload exercises the terminal semantics consumed by the diagnosis
contract across different native control surfaces.  They describe the pinned
applications and configurations under capability-matched scenarios; model,
prompt, tool-description, and orchestration effects remain jointly represented
in each observed execution.

\begin{table}[!htbp]
  \centering
  \caption{The frozen workload exercises blocked, failed, attempted, executed,
  and benign terminal semantics across the pinned target configurations.}
  \label{tab:agent-behavior}
  \footnotesize
  \setlength{\tabcolsep}{5pt}
\begin{tabular*}{\textwidth}{@{\extracolsep{\fill}}lrrrrrrr@{}}
    \toprule
    Target & Runs & Forb. exec. & Forb. att. & Blocked & Auth. fail & Auth. exec. & Benign \\
    \midrule
    TaskWeaver & 15 & 0 & 0 & 10 & 1 & 0 & 4 \\
    Agent Zero & 36 & 6 & 0 & 17 & 2 & 3 & 8 \\
    LangGraph & 36 & 0 & 0 & 0 & 3 & 6 & 27 \\
    Goose & 36 & 5 & 3 & 0 & 3 & 6 & 19 \\
    Microsoft AF & 36 & 0 & 0 & 0 & 3 & 6 & 27 \\
    \bottomrule
  \end{tabular*}
\end{table}

\begin{table}[!htbp]
  \centering
  \caption{Complete paired tracing audit over semantically equivalent
  tracing-on/off executions.  Token entries remain N/A where provider usage
  telemetry was unavailable.}
  \label{tab:rq4-overhead}
  \footnotesize
  \setlength{\tabcolsep}{5pt}
\begin{tabular*}{\textwidth}{@{\extracolsep{\fill}}lcrrrc@{}}
    \toprule
    Target & Valid/selected & Diagnosis (ms) & E2E $\Delta$ & Package $\Delta$ (B) & Token $\Delta$ \\
    \midrule
    Agent Zero & 13/18 & 5.881 & 6.20\% & 53,538 & N/A \\
    Goose & 14/17 & 12.261 & -1.08\% & 112,955 & N/A \\
    LangGraph & 12/15 & 10.893 & 12.76\% & 92,600 & N/A \\
    Microsoft AF & 13/15 & 9.009 & -5.54\% & 56,463 & N/A \\
    TaskWeaver & 11/11 & 6.661 & -0.28\% & 29,849 & N/A \\
    \bottomrule
  \end{tabular*}
\end{table}

\end{document}